\documentclass[11pt]{elsarticle}
\usepackage{bm}
\usepackage[margin=1in,papersize={8.5in,11in}]{geometry}
\usepackage{times,xcolor,hyperref}
\usepackage{array}
\usepackage{amssymb,amsfonts,amsmath,amsthm, mathrsfs,mathtools}
\usepackage{mathabx}
\usepackage{caption}
\usepackage{graphicx}
\usepackage{subcaption}
\usepackage{algorithm}
\usepackage{algpseudocode}
\usepackage{multicol, multirow}
\usepackage{nomencl}
\usepackage{tikz} % for overlayed icons in the parameter catagories table
\usepackage{booktabs, multirow, xcolor}
\makenomenclature
\usepackage{xspace}

\usepackage{subcaption}
\usepackage[most]{tcolorbox}

\newcommand{\bmp}[1]{\begin{minipage}{#1\textwidth}}
\newcommand{\emp}{\end{minipage}}

\DeclareMathOperator{\sinc}{sinc}

\journal{Arxiv}

\begin{document}

\begin{frontmatter}

\title{Uncertainty propagation in auto-regressive random neural network models}

% \author[author]{Jeremy Diamzon}
% \ead{jdiamzon@ucsc.edu}

\author[author]{Janice Adams}
\ead{jaloadam@ucsc.edu }

\author[author]{Daniele Venturi\corref{correspondingAuthor}}
\cortext[correspondingAuthor]{Corresponding author}
\ead{venturi@ucsc.edu}

\address[author]{Department of Applied Mathematics, UC Santa Cruz, Santa Cruz, CA 95064}

\begin{abstract}

We develop analytical and particle-based methods for uncertainty propagation
in random neural network models, where both the inputs and network parameters
are allowed to be random. Building on the piecewise-linear structure of the
Leaky ReLU activation function, we derive a local approximation of the neural
network output with respect to perturbations in both its inputs and parameters.
This approximation is exact for perturbations that preserve the network
activation pattern, and it allows us to compute analytical expressions for the
probability density function and characteristic function of the network output,
together with closed-form approximations for its mean and covariance.
We extend this uncertainty propagation framework to autonomous dynamical
systems whose one-step evolution map is represented by a random neural network.
Repeated application of this map defines an autoregressive model, for which we
derive recursive equations to propagate uncertainty in both the state and
network parameters over time.
These equations explicitly account for the state--parameter cross-covariance
that develops under successive iterations of the network.
Numerical experiments on the Lorenz--63 system and the Kuramoto--Sivashinsky
equation demonstrate accurate uncertainty propagation through the
predictability horizon and the applicability of the proposed framework to
high-dimensional dynamical systems.
\end{abstract}

\end{frontmatter}

\section{Introduction}
\label{sec:intro}

In the past decade, Scientific Machine Learning (SciML) has made significant advances across a wide range of disciplines, including physics, engineering, and medicine \cite{abdar_review_2021, gawlikowski_survey_2023, Panos}. Despite this progress, the development of reliable methods for assessing the uncertainty and trustworthiness of neural network predictions has lagged behind their rapidly increasing capabilities. This gap becomes critical in applications involving nonlinear dynamical systems modeled auto-regressively by neural networks, where small uncertainties in inputs or model parameters can be amplified by the neural network, leading to inaccurate or unstable predictions \cite{Raissi, GK2020, psaros_uncertainty_2023, zou_uncertainty_2025, VenturiSpectral, venturi2018numerical, NNFDE2024}. Uncertainty quantification in deep learning has grown substantially as a field over the past decade \cite{abdar_review_2021, gawlikowski_survey_2023, jospin2022hands, arbel2023primer, goan2020bnn_survey}, with the majority of existing approaches adopting a Bayesian perspective. Well-established methods within this framework include Variational Inference \cite{blundell2015weight}, Monte Carlo Dropout \cite{gal2016dropout}, the Laplace Approximation \cite{mackay1992, ritter2018, immer_improving_2021, daxberger2021},
and Deep Ensembles \cite{lakshminarayanan2017simple}. These methods have proven effective for characterizing model uncertainty in various prediction tasks; however, their application in the \textit{auto-regressive prediction} setting remains largely unexplored, with \cite{shahid_hopcast_nodate} being a notable exception.

 A key property of the Leaky ReLU activation function is that it is piecewise linear, meaning a first-order Taylor expansion of the network output with respect to both the input and the parameters is exact on each linear region. This makes linearization a well-suited tool for analytically characterizing how uncertainty in both the input and parameters propagates through the network. Building on this observation and extending the framework of \cite{diamzon2025uncertainty} to include parameter uncertainty, we derive analytical expressions for the PDF and characteristic function of the network output, and show that the mean and covariance can be computed in closed form via a combined Jacobian of the network with respect to both input and parameters. We extend these results to the auto-regressive setting, where the network is applied repeatedly as a time-stepping map, and develop a particle-based resampling method that remains stable over long time horizons. A naive single-Gaussian propagation, by contrast, becomes unstable. We demonstrate the framework on the Lorenz-63 system and the Kuramoto--Sivashinsky equation. While this paper focuses on Leaky ReLU, the framework readily extends to other activation functions and naturally accommodates higher-order expansions.

This paper is organized as follows. In Section~\ref{sec:background}, we introduce the multi-layer perceptron architecture and notation. In Section~\ref{sec:linearized_net}, we derive the linearized approximation of the network output with respect to both the input and the parameters, and discuss related work. In Section~\ref{sec:pdf_approximation}, we derive analytical expressions for the PDF and characteristic function of the network output. In Section~\ref{sec:Statistical_Moments}, we derive closed-form approximations for the mean and covariance of the network output given the linearized model. In Section~\ref{sec:autoregressive}, we extend these methods to the auto-regressive setting and present propagation algorithms. In Section~\ref{sec:numerical_experiments}, we present numerical experiments on the Lorenz-63 system and the Kuramoto--Sivashinsky equation. Section~\ref{sec:conclusion} concludes.

\begin{figure}[t]
    \centering
    \includegraphics[width=0.5\linewidth]{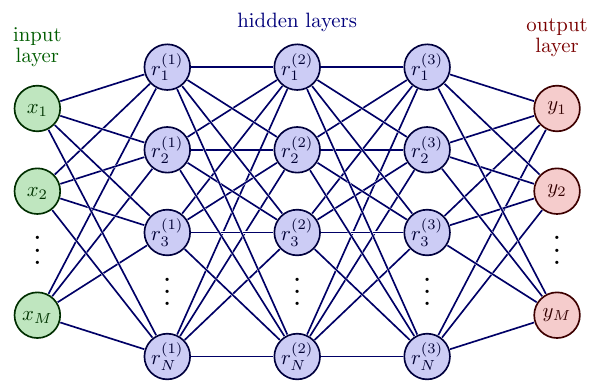}
    \caption{A diagram of a fully connected feed-forward neural network, also known as a multi-layer perceptron (MLP). The input and output in this diagram are vectors of dimension $M$, and each of the three hidden layers have dimension $N$. 
%    This diagram was generated via LaTeX TikZ library \cite{neutelings2021tikz}. 
    }
    \label{fig:MLP_diagram}
\end{figure}

\section{Neural Network Model}
\label{sec:background}

Let $\bm y = \bm F_{\bm \theta}(\bm x)$ represent a fully-connected feed-forward neural network that maps input $\bm x$ to output $\bm y$, with parameters $\bm \theta$ and $H$ hidden layers, as in Figure \ref{fig:MLP_diagram}. We set $\bm x, \bm y \in\mathbb{R}^M$ and each hidden layer to have dimension $N$. The mapping between each layer is given by an affine transformation and a nonlinear activation function $\phi$: 
\begin{align}
    \bm{r}_1 &= \phi\left(\bm W_1 \bm x + \bm b_1\right) \nonumber \\
    \bm r_i &= \phi\left(\bm W_i \bm r_{i-1} +\bm b_i\right) , \qquad i=2,\dots,H \nonumber \\
    \bm y &= \bm W_{H+1}\bm r_H + \bm b_{H+1}, 
    \label{eq:nn_layers}
\end{align}
where the parameters consist of weights and biases $\bm \theta = \{\bm W_i, \bm b_i\}_{i=1}^{H+1}$. The dimensions of the parameters are as follows: 
\begin{alignat}{2}
    & \bm W_1 \in\mathbb{R}^{N\times M}, \qquad && \bm b_1\in\mathbb{R}^N \nonumber \\
    & \bm W_i\in\mathbb{R}^{N\times N}, \qquad && \bm b_i\in\mathbb{R}^N, \qquad i=2,\dots,H \nonumber \\
    & \bm W_{H+1}\in\mathbb{R}^{M\times N}, \qquad && \bm b_{H+1}\in\mathbb{R}^M.
\end{alignat}
As previously stated, we set the activation functions for all layers to be Leaky ReLU, defined component-wise as
\begin{equation}
    \phi(x) = \begin{cases}
  x & \text{if } x > 0 \\
  \alpha x & \text{if } x \leq 0
\end{cases}
\end{equation}
for constant $\alpha$, defaulting to $\alpha = 0.01$.

\section{Linearization of the Neural Network Output}
\label{sec:linearized_net}

A useful approach for analyzing the input--output behavior of a neural
network is to linearize its output with respect to the input. Such
linearizations, typically obtained through a Taylor expansion, have been used
to aid interpretability and explainability \citep{montavon2017deep_taylor, bach2015lrp}, add physics-informed structure \citep{zhu2022nnpoly, huang2024physics_taylor, taylor_maps2019}, and improve inference accuracy or efficiency \citep{zwerschke2024taylor, rueckauer2019linear_video, xiao2023hope}. Expansions with respect to the parameters are also used to study optimization and generalization, e.g. via the neural tangent kernel \citep{balduzzi2017neural_taylor, lee2019wide, jacot2018ntk}, and form the basis of Laplace approximations for Bayesian uncertainty quantification \citep{mackay1992, ritter2018, daxberger2021}. 
A smaller body of work addresses uncertainty propagation directly, via
linearization with respect to the input \citep{diamzon2025uncertainty, titensky2018ekf, jungmann2025analytical, petersen2024stable} or the parameters \citep{immer_improving_2021}.

In particular, in \cite{diamzon2025uncertainty} we observed that linearization
is particularly effective for neural networks with leaky ReLU activation 
functions. Indeed, because leaky ReLU is piecewise
linear, the local linearization is exact as long as the perturbation does not
cause a change in the active branches of the activation functions. More
generally, even when activation branches change, the linearized representation
was found to provide accurate statistical predictions for relatively large
input perturbations. This behavior can be attributed to the way perturbations
propagate through the network.
\citet{titensky2018ekf}, \citet{jungmann2025analytical}, and
\citet{petersen2024stable} similarly linearize the network output with
respect to the input alone, while keeping the network parameters fixed.
\citet{immer_improving_2021} instead linearize the network output with
respect to the parameters for Laplace-approximate Bayesian inference.

Building upon our recent results \cite{diamzon2025uncertainty}, here we consider
a first-order approximation of the neural network defined in
Section~\ref{sec:background} with respect to both the input $\bm x$ and the
parameters $\bm\theta$. Specifically, we regard
$\bm F_{\bm\theta}(\bm x)$ as a function of both arguments and consider
perturbations $\Delta\bm x$ and $\Delta\bm\theta$. The corresponding
first-order approximation of the network output is
\begin{equation}
\bm F_{\bm\theta+\Delta\bm\theta}
\left(\bm x+\Delta\bm x\right)
\approx
\bm F_{\bm\theta}(\bm x)
+\bm J_{\bm x}(\bm x,\bm \theta)\Delta\bm x
+\bm J_{\bm\theta}(\bm x,\bm \theta)\Delta\bm\theta,
\label{linNet}
\end{equation}
where
\begin{align}
\bm J_{\bm x}(\bm x,\bm \theta)
=
\frac{\partial \bm F_{\bm\theta}(\bm x)}{\partial \bm x},
\qquad
\bm J_{\bm\theta}(\bm x,\bm \theta)
=
\frac{\partial \bm F_{\bm\theta}(\bm x)}{\partial \bm\theta},
\label{jacobians}
\end{align}
 are the Jacobians of the network with respect to the input and the parameters, respectively. 

\section{PDF of the Neural Network Output}
\label{sec:pdf_approximation}
%In this section we derive an approximation for the probability density function (PDF) and the characteristic function of the neural net output, given our linearization of an MLP in Section \ref{sec:linearized_net}. 

We first derive a general expression for the output PDF of an MLP. We then specialize this expression to the linearized neural network \eqref{linNet}. Finally, we derive explicit forms of the resulting approximate PDF for several classes of perturbations, including independent perturbations with general distributions, Gaussian perturbations, and uniform perturbations.

To this end, let $\bm x$ and $\bm\theta$ be random vectors representing the input and
parameters of the network, respectively, and let $\bm y$ denote the resulting
random output vector. Since the network defines a deterministic mapping for fixed
$\bm x$ and $\bm\theta$, the conditional PDF of $\bm y$ given $\bm x$ and $\bm \theta$ is
\begin{equation}
p\left(\bm y \mid \bm x,\bm\theta\right)
=
\delta\left(\bm y-\bm F_{\bm\theta}(\bm x)\right),
\label{eq:p_y_given_x_theta}
\end{equation}
where $\delta(\cdot)$ denotes the Dirac delta function.
Marginalizing over the joint distribution of $\bm x$ and $\bm\theta$ yields
the output PDF
\begin{align}
p(\bm y)
&=
\int\int
p\left(\bm y\mid\bm x,\bm\theta\right)
p(\bm x,\bm\theta)
\,d\bm x\,d\bm\theta
\nonumber\\
&=
\int\int
\delta\left(\bm y-\bm F_{\bm\theta}(\bm x)\right)
p(\bm x,\bm\theta)
\,d\bm x\,d\bm\theta,
\label{eq:PDF_g_general_dirac}
\end{align}
where $p(\bm x,\bm\theta)$ denotes the joint PDF of the input and network
parameters, and the integrals are taken over its support.
As is well-known, the multivariate Dirac delta can be represented in Fourier form as
\begin{equation}
\delta\left(\bm y-\bm F_{\bm\theta}(\bm x)\right)
=
\frac{1}{(2\pi)^M}
\int_{\mathbb R^M}
e^{i\bm a^\top\left[\bm y-\bm F_{\bm\theta}(\bm x)\right]}
\,d\bm a.
\end{equation}
Substituting this representation into
\eqref{eq:PDF_g_general_dirac} yields the following general (exact) expression for
the PDF of the network output:
\begin{equation}
p(\bm y)
=
\frac{1}{(2\pi)^M}
\int_{\mathbb R^M}
\int\int
e^{i\bm a^\top\left[\bm y-\bm F_{\bm\theta}(\bm x)\right]}
p(\bm x,\bm\theta)
\,d\bm x\,d\bm\theta\,d\bm a.
\label{eq:pdf_g_general}
\end{equation}

\subsection{PDF of the Linearized Neural Network Model}
We decompose the random vectors $\bm x$ and $\bm\theta$ into their means and
corresponding perturbations,
\begin{align}
\bm x &= \bar{\bm x}+\Delta\bm x, \nonumber\\
\bm\theta &= \bar{\bm\theta}+\Delta\bm\theta,
\label{eq:mean_and_pertubations}
\end{align}
where $(\Delta\bm x,\Delta\bm\theta)$ has joint PDF
$p(\Delta\bm x,\Delta\bm\theta)$. Substituting these decompositions into
\eqref{eq:pdf_g_general} allows us to express the PDF of $\bm y$ in terms of the
perturbations as
\begin{equation}
p(\bm y)
=
\frac{1}{(2\pi)^M}
\int_{\mathbb R^M}
\int\int
e^{i\bm a^\top
\left[
\bm y-
\bm F_{\bar{\bm\theta}+\Delta\bm\theta}
(\bar{\bm x}+\Delta\bm x)
\right]}
p(\Delta\bm x,\Delta\bm\theta)
\,d\Delta\bm x\,d\Delta\bm\theta\,d\bm a.
\label{eq:pdf_g_general_zero_mean}
\end{equation}
We now consider the following first-order approximation of the network output
\begin{equation}
\bm F_{\bar{\bm\theta}+\Delta\bm\theta}
\left(\bar{\bm x}+\Delta\bm x\right)
\approx
\bm F_{\bar{\bm\theta}}(\bar{\bm x})
+
\bm J(\bar{\bm x},\bar{\bm\theta})
\begin{bmatrix}
\Delta\bm x\\
\Delta\bm\theta
\end{bmatrix},
\label{FOA}
\end{equation}
where
\begin{equation}
\bm J(\bar{\bm x},\bar{\bm\theta})
=
\begin{bmatrix}
\bm J_{\bm x}(\bar{\bm x},\bar{\bm\theta}) &
\bm J_{\bm\theta}(\bar{\bm x},\bar{\bm\theta})
\end{bmatrix},
\end{equation}
and the Jacobians $\bm J_{\bm x}$ and $\bm J_{\bm\theta}$ are defined
in \eqref{jacobians}. Substituting \eqref{FOA} into
\eqref{eq:pdf_g_general_zero_mean} yields
\begin{align}
p(\bm y)
\approx {}&
\frac{1}{(2\pi)^M}
\int_{\mathbb R^M}
\int\int
\exp\left(i\bm a^\top
\left[
\bm y-\bm F_{\bar{\bm\theta}}(\bar{\bm x})
-\bm J
\begin{bmatrix}
\Delta\bm x\\
\Delta\bm\theta
\end{bmatrix}
\right]\right)
p(\Delta\bm x,\Delta\bm\theta)
\,d\Delta\bm x\,d\Delta\bm\theta\,d\bm a
\nonumber\\
={}&
\frac{1}{(2\pi)^M}
\int_{\mathbb R^M}
\exp\left(i\bm a^\top
\left[
\bm y-\bm F_{\bar{\bm\theta}}(\bar{\bm x})
\right]\right)
\left(
\int\int
\exp\left(-i\bm a^\top\bm J
\begin{bmatrix}
\Delta\bm x\\
\Delta\bm\theta
\end{bmatrix}\right)
p(\Delta\bm x,\Delta\bm\theta)
\,d\Delta\bm x\,d\Delta\bm\theta
\right)d\bm a.
\label{eq:pdf_g_approx}
\end{align}
The inner integral with respect to $\Delta\bm x$ and 
$\Delta\bm\theta$ is the joint characteristic function of
$(\Delta\bm x,\Delta\bm\theta)$ evaluated at
\begin{equation}
-\bm J^\top\bm a
=
\left(
-\bm J_{\bm x}^\top\bm a,
-\bm J_{\bm\theta}^\top\bm a
\right).
\end{equation}
Denoting the joint characteristic function by
$\varphi_{\Delta\bm x,\Delta\bm\theta}$, we can rewrite
\eqref{eq:pdf_g_approx} compactly as
\begin{equation}
p(\bm y)
\approx
\frac{1}{(2\pi)^M}
\int_{\mathbb R^M}
e^{i\bm a^\top
\left[
\bm y-\bm F_{\bar{\bm\theta}}(\bar{\bm x})
\right]}
\varphi_{\Delta\bm x,\Delta\bm\theta}
\left(-\bm J^\top\bm a\right)
\,d\bm a.
\label{eq:pdf_g_approx_in_terms_of_char_eq}
\end{equation}
Equation~\eqref{eq:pdf_g_approx_in_terms_of_char_eq} shows that, 
under the first-order approximation \eqref{FOA} the PDF of the output 
is completely determined by the joint characteristic function
of $(\Delta\bm x,\Delta\bm\theta)$. In particular, setting
$\bm\omega=-\bm a$ allows us to identify the approximate characteristic
function of the output $\bm y$ as
\begin{equation}
\varphi_{\bm y}(\bm\omega)
\approx
e^{i\bm\omega^\top\bm F_{\bar{\bm\theta}}(\bar{\bm x})}
\varphi_{\Delta\bm x,\Delta\bm\theta}
\left(\bm J^\top\bm\omega\right).
\label{eq:char_g_approx}
\end{equation}

\subsection{Statistical Models}
We now specialize the PDF approximation
\eqref{eq:pdf_g_approx_in_terms_of_char_eq} to three statistical models for
the input and parameter perturbations, namely correlated multivariate Gaussian
perturbations, statistically independent perturbations with general
marginals, and independent uniform perturbations.

\subsubsection{Multivariate Gaussian Perturbations}
\label{sec:Multivariate Gaussian}
Assume that the input and parameter perturbations are jointly Gaussian,
\begin{equation}
\begin{bmatrix}
\Delta\bm x\\
\Delta\bm\theta
\end{bmatrix}
\sim
\mathcal{N}\left(\bm 0,\bm\Sigma_{\Delta}\right),
\end{equation}
where $\bm\Sigma_{\Delta}$ is the covariance matrix of 
the stacked perturbation vector
$\left[\Delta\bm x^\top,\Delta\bm\theta^\top\right]^\top$.
The characteristic function of $(\Delta \bm x, \Delta \bm \theta)$ is given by 
\begin{equation}
    \varphi_{\Delta \bm x, \Delta \bm \theta}\left(\bm u\right) = e^{-\frac{1}{2}\bm u^\top \bm \Sigma_{\Delta} \bm u}. 
\end{equation}
Evaluating $ \varphi_{\Delta \bm x, \Delta \bm \theta}$ at $\bm u = -\bm J^\top  \bm a$, where $\bm J$ is again the combined Jacobian, gives
\begin{equation}
    \varphi_{\Delta \bm x, \Delta \bm \theta}\left(-\bm J_{\bm x}^\top \bm a, -\bm J_{\bm \theta}^\top \bm a\right) = e^{-\frac{1}{2} \bm a^\top\bm J \bm \Sigma_{\Delta} \bm J^\top  \bm a}.
\end{equation}
Substituting this into \eqref{eq:pdf_g_approx_in_terms_of_char_eq} gives
\begin{equation}
    p(\bm y) \approx \frac{1}{(2\pi)^{M}} 
\int_{\mathbb{R}^{M}}
\exp\left(i\bm a^\top \left(\bm y-\bm F_{\bar{\bm \theta}}(\bar{\bm x}) \right)-\frac{1}{2} \bm a^\top \bm J \bm \Sigma_{\Delta}\bm J^\top \bm a \right) d\bm a. 
\end{equation}
This is the inverse Fourier transform of a Gaussian characteristic function, 
so we have that
\begin{equation}
    \bm y \sim \mathcal{N} \left(\bm F_{\bar{\bm \theta}}(\bar{\bm x}), \bm J \bm \Sigma_{\Delta}\bm J^\top \right)
\end{equation}
under the linearized approximation.

\subsubsection{Statistically Independent Perturbations}

Assume that the random perturbation vectors $\Delta\bm x$ and
$\Delta\bm\theta$ are statistically independent, i.e.,
\begin{equation}
p(\Delta\bm x,\Delta\bm\theta)
=
p(\Delta\bm x)\,p(\Delta\bm\theta).
\end{equation}
The joint characteristic function then factorizes as
\begin{equation}
\varphi_{\Delta\bm x,\Delta\bm\theta}
\left(-\bm J^\top\bm a\right)
=
\varphi_{\Delta\bm x}
\left(-\bm J_{\bm x}^\top\bm a\right)
\varphi_{\Delta\bm\theta}
\left(-\bm J_{\bm\theta}^\top\bm a\right).
\end{equation}
Substituting this factorization into
\eqref{eq:pdf_g_approx_in_terms_of_char_eq} and
\eqref{eq:char_g_approx} yields the corresponding expressions for
$p(\bm y)$ and $\varphi_{\bm y}$.
This factorization can be extended under additional independence assumptions
on the components of $\Delta\bm x$ and $\Delta\bm\theta$. For example,
suppose that the parameter perturbations associated with the different network
layers, $\{\Delta\bm\theta_1, \ldots, \Delta\bm\theta_{H+1}\}$, are 
mutually independent and independent of the input perturbation $\Delta\bm x$. 
The joint PDF then factorizes as
\begin{equation}
p(\Delta\bm x,\Delta\bm\theta_1,\ldots,\Delta\bm\theta_{H+1})
=
p(\Delta\bm x)
\prod_{i=1}^{H+1}p(\Delta\bm\theta_i),
\end{equation}
and the corresponding joint characteristic function becomes
\begin{equation}
\varphi_{\Delta\bm x,\Delta\bm\theta}
\left(-\bm J^\top\bm a\right)
=
\varphi_{\Delta\bm x}
\left(-\bm J_{\bm x}^\top\bm a\right)
\prod_{i=1}^{H+1}
\varphi_{\Delta\bm\theta_i}
\left(-\bm J_{\bm\theta_i}^\top\bm a\right).
\end{equation}

\subsubsection{Independent Uniform Perturbations}

Assume that the entries of the stacked perturbation vector
$ \Delta\bm z = \begin{bmatrix}
\Delta\bm x; 
\Delta\bm\theta
\end{bmatrix} $ are statistically independent and uniformly distributed as
\begin{equation}
\Delta z_k \sim \mathcal{U}(-\beta_k,\beta_k),
\qquad k=1,\ldots, d_{\Delta},
\end{equation}
where $d_{\Delta}$ is the dimension of the stacked perturbation vector.
For a centered uniform random variable
$X\sim\mathcal{U}(-\beta,\beta)$, the characteristic function is
\begin{equation}
\varphi_X(u)
=
\frac{\sin(\beta u)}{\beta u}
=
\sinc(\beta u).
\end{equation}
Therefore, by independence,
\begin{equation}
\varphi_{\Delta\bm x,\Delta\bm\theta}(\bm u)
=
\prod_{k=1}^{d_{\Delta}}
\sinc(\beta_k u_k).
\label{eq:char_eq_uniform}
\end{equation}
Evaluating this characteristic function at
$\bm u=-\bm J^\top\bm a$ and using the fact that $\sinc(\cdot)$ is an
even function, we obtain
\begin{equation}
\varphi_{\Delta\bm x,\Delta\bm\theta}
\left(-\bm J^\top\bm a\right)
=
\prod_{k=1}^{d_{\Delta}}
\sinc\left(
\beta_k[\bm J^\top\bm a]_k
\right).
\end{equation}
Substituting this expression into
\eqref{eq:pdf_g_approx_in_terms_of_char_eq} yields
\begin{equation}
p(\bm y)
\approx
\frac{1}{(2\pi)^M}
\int_{\mathbb R^M}
e^{i\bm a^\top
\left[
\bm y-\bm F_{\bar{\bm\theta}}(\bar{\bm x})
\right]}
\prod_{k=1}^{d_{\Delta}}
\sinc\left(
\beta_k[\bm J^\top\bm a]_k
\right)
\,d\bm a.
\end{equation}

\subsection{Connection to Bayesian Neural Networks (BNNs)}
The probabilistic formulation developed above is closely related to Bayesian
Neural Networks (BNNs). In a standard BNN, uncertainty in the network
parameters is represented through a posterior distribution
$p(\bm\theta\mid\bm D)$ conditioned on the training data $\bm D$. Given a new
input $\bm x^*$, predictions are obtained from the posterior predictive
distribution
\begin{equation}
p\left(\bm y^* \mid \bm x^*,\bm D\right)
=
\int
p\left(\bm y^* \mid \bm F_{\bm\theta}(\bm x^*)\right)
p(\bm\theta\mid\bm D)
\,d\bm\theta.
\label{eq:posterior_predictive_distribuion}
\end{equation}
Thus, both BNNs and the framework developed in this paper propagate
uncertainty in the network parameters through the nonlinear mapping
$\bm F_{\bm\theta}$. There are, however, several important differences
between the two formulations.
First, our framework does not prescribe how the probability distribution of
the network parameters is obtained. In a BNN, $p(\bm\theta\mid\bm D)$
is inferred from the training data using a Bayesian inference procedure.
Since the exact posterior is generally intractable for neural networks,
various approximations are commonly used, including variational inference \cite{blundell2015weight}, Laplace approximations \cite{mackay1992, ritter2018, daxberger2021}, and ensemble-based approaches \cite{lakshminarayanan2017simple}.
In this paper, by contrast, a probability distribution describing uncertainty in
$\bm\theta$ is taken as given and subsequently propagated through the network.
Consequently, the analysis is independent of the particular procedure used
to construct this distribution. In the numerical experiments discussed below, 
parameter uncertainty is estimated from network parameters
collected over multiple training epochs.
Second, we assume that the neural network defines a deterministic mapping
once its input and parameters are specified. Accordingly, the conditional
output PDF is represented by
\begin{equation}
p\left(\bm y\mid\bm x,\bm\theta\right)
=
\delta\left(\bm y-\bm F_{\bm\theta}(\bm x)\right).
\end{equation}
This differs from BNN formulations in which the likelihood explicitly
includes an observation-noise or output-noise model
\cite{jospin2022hands, arbel2023primer}. Our formulation therefore isolates
the uncertainty induced by uncertain inputs and uncertain network parameters,
rather than introducing an additional stochastic output model.
Third, and most importantly for the nonlinear dynamical systems applications 
considered in this paper, we treat both the network input $\bm x$ and 
the network parameters $\bm\theta$ as random. The corresponding 
output distribution is therefore obtained by marginalizing over 
their joint distribution,
\begin{equation}
p(\bm y)
=
\int\int
\delta\left(\bm y-\bm F_{\bm\theta}(\bm x)\right)
p(\bm x,\bm\theta)
\,d\bm x\,d\bm\theta.
\end{equation}
The standard posterior predictive distribution
\eqref{eq:posterior_predictive_distribuion} is recovered as a particular case
when the input $\bm x=\bm x^*$ is treated as deterministic and uncertainty is
introduced only through the network parameters. Allowing $\bm x$ to be random
is especially important in auto-regressive applications, because the output
distribution produced at one time step becomes the uncertain input
distribution at the next. Hence, uncertainty must be propagated recursively
through successive applications of the neural network.
Another important difference lies in how the resulting integrals are evaluated. BNN posterior predictive distributions are commonly approximated by sampling from the parameter posterior or by using an approximate posterior representation; a linearized alternative that avoids this sampling in the parameter-only case is the ``GLM predictive'' \citep{immer_improving_2021}. Here, after applying the first-order approximation \eqref{FOA}, the propagation of uncertainty can instead be expressed directly in terms of the characteristic function of the joint input and parameter perturbations. As shown in \eqref{eq:pdf_g_approx_in_terms_of_char_eq}, this yields an analytical representation of the approximate output PDF without requiring repeated sampling of the neural network.

\section{Statistical Moments}
\label{sec:Statistical_Moments}
We derive closed-form approximations for the mean and covariance of the neural
network output $\bm y$ under the linearized model \eqref{linNet}. Unlike the
PDF and characteristic-function expressions discussed in
Section~\ref{sec:pdf_approximation}, these moment equations remain tractable
in the auto-regressive setting discussed below, even when the 
joint density of the state and parameter perturbations is no longer 
available in closed form.
A key advantage of the linearized model is that it also preserves important
classes of probability distributions under affine transformations. In
particular, Gaussian distributions are closed under affine maps. Therefore,
if the joint perturbation $[\Delta\bm x;\Delta\bm\theta]$ is Gaussian, the
linearized network output is also exactly Gaussian, with mean and covariance
given directly by the expressions derived below.

Recall the first-order approximation \eqref{FOA} of the neural network 
output. By construction, the perturbations $\Delta\bm x$ 
and $\Delta\bm\theta$ have zero mean,
\begin{equation}
\mathbb E[\Delta\bm x]=\bm 0,
\qquad
\mathbb E[\Delta\bm\theta]=\bm 0.
\end{equation}
It follows that the mean of the network output can be approximated as
\begin{align}
\mathbb E[\bm y]
&=
\mathbb E\left[
\bm F_{\bar{\bm\theta}+\Delta\bm\theta}
(\bar{\bm x}+\Delta\bm x)
\right]
\nonumber\\
&\approx
\bm F_{\bar{\bm\theta}}(\bar{\bm x})
+
\bm J
\mathbb E
\begin{bmatrix}
\Delta\bm x\\
\Delta\bm\theta
\end{bmatrix}
\nonumber\\
&=
\bm F_{\bar{\bm\theta}}(\bar{\bm x}).
\label{eq:mean_y}
\end{align}
Thus, under the first-order approximation, the mean output is given by the
network evaluated at the mean input and mean parameter values.
The covariance of the network output is defined as
\begin{equation}
\bm\Sigma_{\bm y}
=
\mathbb E\left[
(\bm y-\mathbb E[\bm y])
(\bm y-\mathbb E[\bm y])^\top
\right].
\end{equation}
Using the first-order approximation and \eqref{eq:mean_y}, we have
\begin{align}
\bm\Sigma_{\bm y}
&\approx
\mathbb E\left[
\left(
\bm J
\begin{bmatrix}
\Delta\bm x\\
\Delta\bm\theta
\end{bmatrix}
\right)
\left(
\bm J
\begin{bmatrix}
\Delta\bm x\\
\Delta\bm\theta
\end{bmatrix}
\right)^\top
\right]
\nonumber\\
&=
\bm J
\mathbb E\left[
\begin{bmatrix}
\Delta\bm x\\
\Delta\bm\theta
\end{bmatrix}
\begin{bmatrix}
\Delta\bm x\\
\Delta\bm\theta
\end{bmatrix}^{\!\top}
\right]
\bm J^\top
\nonumber\\
&=
\bm J\bm\Sigma_{\Delta}\bm J^\top,
\label{eq:Sigma_y}
\end{align}
where the joint covariance matrix of the input and parameter perturbations
can be written in block form as
\begin{equation}
\bm\Sigma_{\Delta}
=
\begin{bmatrix}
\bm\Sigma_{\Delta\bm x}
&
\bm\Sigma_{\Delta\bm x,\Delta\bm\theta}
\\
\bm\Sigma_{\Delta\bm\theta,\Delta\bm x}
&
\bm\Sigma_{\Delta\bm\theta}
\end{bmatrix}.
\label{eq:joint_coF_blocked}
\end{equation}
Since $\Delta\bm x=\bm x-\bar{\bm x}$, its covariance is identical to that
of $\bm x$, so that
$\bm\Sigma_{\Delta\bm x}=\bm\Sigma_{\bm x}$.
Substituting the block representation
\eqref{eq:joint_coF_blocked} into \eqref{eq:Sigma_y} gives
\begin{align}
\bm\Sigma_{\bm y}
\approx {}
\bm J_{\bm x}\bm\Sigma_{\bm x}\bm J_{\bm x}^\top
+
\bm J_{\bm\theta}\bm\Sigma_{\Delta\bm\theta}
\bm J_{\bm\theta}^\top
+
\bm J_{\bm x}
\bm\Sigma_{\Delta\bm x,\Delta\bm\theta}
\bm J_{\bm\theta}^\top
+
\bm J_{\bm\theta}
\bm\Sigma_{\Delta\bm\theta,\Delta\bm x}
\bm J_{\bm x}^\top.
\label{eq:Sigma_y_expanded}
\end{align}
The first term represents the contribution of input uncertainty, the second
the contribution of parameter uncertainty, and the last two terms account
for cross-covariance between the input and parameter perturbations.
If $\Delta\bm x$ and $\Delta\bm\theta$ are uncorrelated, then
\begin{equation}
\bm\Sigma_{\Delta\bm x,\Delta\bm\theta}
=
\bm\Sigma_{\Delta\bm\theta,\Delta\bm x}
=
\bm 0,
\end{equation}
and \eqref{eq:Sigma_y_expanded} simplifies to
\begin{equation}
\bm\Sigma_{\bm y}
\approx
\bm J_{\bm x}\bm\Sigma_{\bm x}\bm J_{\bm x}^\top
+
\bm J_{\bm\theta}\bm\Sigma_{\Delta\bm\theta}
\bm J_{\bm\theta}^\top.
\end{equation}

\section{Auto-Regressive Neural Networks}
\label{sec:autoregressive}

Auto-regressive neural networks are models that predict the next state of a
sequence based on its previous states. They are widely used for modeling
time-series data in applications such as weather and climate forecasting,
electronic health records, and financial prediction~\cite{lim2021time}.
A basic one-step auto-regressive model based on the MLP described in
Section~\ref{sec:background} maps the state $\bm x_t$ at time $t$ to the
state at the next time step via
\begin{equation}
\bm x_{t+1} = \bm F_{\bm\theta}(\bm x_t).
\label{autoreg}
\end{equation}
The network is then applied recursively, with each predicted state used as
the input at the next time step, to generate a time-series trajectory.
Several extensions of the basic auto-regressive architecture have been
developed to improve predictive performance, including lagged-input window
models and recurrent architectures such as Long Short-Term Memory (LSTM)
networks~\cite{hochreiter1997lstm}. For a broader discussion of
surrogate-modeling approaches for uncertainty quantification in time-dependent
nonlinear dynamical systems, including polynomial chaos expansions and
nonlinear autoregressive models with exogenous inputs (NARX), we refer to
\citet{marelli2025surrogate}.

In this section, we aim to characterize the evolution of the state PDF
$p(\bm x_t)$, given an initial joint distribution of the state
and network parameters $p(\bm x_t,\bm\theta)$. The propagation of the PDF
through successive applications of the neural network can be expressed as
the following recursive analogue of \eqref{eq:PDF_g_general_dirac}
\begin{equation}
p(\bm x_{t+1})
=
\int\int
p(\bm x_{t+1}\mid\bm x_t,\bm\theta)
\,p(\bm x_t,\bm\theta)
\,d\bm x_t\,d\bm\theta.
\label{eq:autoregressive_pdf}
\end{equation}
Note that when the state and network parameters are correlated,
$p(\bm x_t,\bm\theta)$ cannot be factorized as
$p(\bm x_t)p(\bm\theta)$. Consequently, the evolution of the state PDF
cannot, in general, be expressed solely in terms of a transition density
$p(\bm x_{t+1}\mid\bm x_t)$ without accounting for the joint dependence
between $\bm x_t$ and $\bm\theta$.
This contrasts with standard nonlinear auto-regressive Markov processes, such
as those commonly used in particle filtering and recursive Bayesian state
estimation~\cite{wills2023smc}, where the dynamics are modeled as
\begin{equation}
\bm x_{k+1}
=
\bm G(\bm x_k)+\bm\xi_k,
\end{equation}
and $\{\bm\xi_k\}$ is a sequence of i.i.d.\ random vectors. Such processes
are Markovian in the state alone, meaning that the conditional distribution
of the next state depends only on the current state. Consequently, the state
PDF can be propagated using the transition density
$p(\bm x_{t+1}\mid\bm x_t)$ as
\begin{equation}
p(\bm x_{t+1})
=
\int
p(\bm x_{t+1}\mid\bm x_t)
\,p(\bm x_t)
\,d\bm x_t.
\label{eq:state_tranisition_density}
\end{equation}
The Markov property can be recovered by augmenting the state vector to include
the network parameters, i.e., $\bm z_t=(\bm x_t,\bm\theta)$. The augmented
state evolves according to the map
\begin{equation}
\bm z_{t+1}
=
\begin{bmatrix}
\bm F_{\bm\theta}(\bm x_t)\\
\bm\theta
\end{bmatrix},
\end{equation}
which defines a Markov process with zero process noise.

Our method also differs from classical particle filtering in the role of
observations. In particle filtering, the goal is to recursively estimate 
the posterior distribution $p(\bm x_k\mid\bm y_1,\ldots,\bm y_k)$ by 
alternating between a prediction step, in which the current state distribution is propagated through the transition model, and an update step, 
in which the predicted distribution is conditioned on the new 
measurement $\bm y_k$ using Bayes' rule. In our
setting, there are no measurements and therefore no update step. 
The goal is therefore pure forecasting: a given initial joint distribution
$p(\bm x_0,\bm\theta)$ 
is propagated forward without assimilating any observations.
A closely related augmented-state prediction framework was recently proposed
in \citet{kuang2025adf} within a full Bayesian filtering setting, tested on a stochastic Lorenz system. Unlike the pure-forecasting setting considered
here, this formulation includes measurement-update steps and treats the
network parameters as fixed rather than uncertain.
From the perspective of classical particle filtering, our method can be interpreted
as an approximation to the prediction step of a Bayesian filter defined on the
augmented state space $\bm z_t$. Recovering
the Markov property in $\bm z_t$, however, does not make the multi-step
evolution of $p(\bm x_t,\bm\theta)$ tractable. Although linearization of
$\bm F_{\bm\theta}$ provides a tractable approximation for one-step
propagation, the same realization of $\bm\theta$ is retained throughout the
forecast, causing $\bm x_t$ and $\bm\theta$ to become correlated. Consequently,
$p(\bm x_t,\bm\theta)$, and hence $p(\bm x_t)$, cannot in general be propagated
in closed form over multiple steps. Hence, additional
approximations are required to characterize the evolution of
$p(\bm x_t)$ in the auto-regressive setting. We discuss these approximations
next.

\subsection{Statistical Moments}
\label{sec:Auto-regressive_Statistical_Moments}
As before, we assume zero-mean perturbations in the state and network
parameters, i.e.,
$\mathbb{E}[\Delta\bm x_t]=\bm 0$ and
$\mathbb{E}[\Delta\bm\theta]=\bm 0$. We first apply the linearization
introduced in Section~\ref{sec:linearized_net} to the auto-regressive mapping
\eqref{autoreg}. Writing
$\bm x_t=\bar{\bm x}_t+\Delta\bm x_t$ and
$\bm\theta=\bar{\bm\theta}+\Delta\bm\theta$, we obtain
\begin{align}
\bm x_{t+1}
&=
\bm F_{\bar{\bm\theta}+\Delta\bm\theta}
(\bar{\bm x}_t+\Delta\bm x_t)
\nonumber\\
&\approx
\bar{\bm x}_{t+1}
+
\bm J_t
\begin{bmatrix}
\Delta\bm x_t\\
\Delta\bm\theta
\end{bmatrix},
\end{align}
where
\begin{equation}
\bar{\bm x}_{t+1}
=
\bm F_{\bar{\bm\theta}}(\bar{\bm x}_t),
\end{equation}
and
$\bm J_t=
[\bm J_{\bm x_t}\ \bm J_{\bm\theta_t}]$
is the combined Jacobian evaluated at
$(\bar{\bm x}_t,\bar{\bm\theta})$.
The auto-regressive linearization can now be substituted directly into
\eqref{eq:mean_y}, \eqref{eq:Sigma_y}, \eqref{eq:joint_coF_blocked}, and
\eqref{eq:Sigma_y_expanded} to obtain recursive approximations for the
statistical moments of $\bm x_{t+1}$. As in \eqref{eq:mean_y}, the mean is
approximated by the unperturbed network output,
\begin{equation}
\mathbb{E}\left[\bm x_{t+1}\right]
\approx
\bar{\bm x}_{t+1}.
\label{eq:mean_y_update}
\end{equation}
Similarly, the covariance of $\bm x_{t+1}$, denoted by
$\bm\Sigma_{t+1}$, is approximated as
\begin{equation}
\bm\Sigma_{t+1}
\approx
\bm J_t\bm\Sigma_{\Delta_t}\bm J_t^\top,
\label{eq:Sigma_y_update}
\end{equation}
where the joint covariance matrix of the state and parameter perturbations
at step $t$ is
\begin{equation}
\bm\Sigma_{\Delta_t}
=
\begin{bmatrix}
\bm\Sigma_t
&
\bm\Sigma_{\Delta\bm x_t,\Delta\bm\theta}
\\
\bm\Sigma_{\Delta\bm\theta,\Delta\bm x_t}
&
\bm\Sigma_{\Delta\bm\theta}
\end{bmatrix}.
\label{eq:Sigma_full_update}
\end{equation}
Expanding \eqref{eq:Sigma_y_update} gives
\begin{align}
\bm\Sigma_{t+1}
\approx {}&
\bm J_{\bm x_t}\bm\Sigma_t\bm J_{\bm x_t}^\top
+
\bm J_{\bm\theta_t}\bm\Sigma_{\Delta\bm\theta}
\bm J_{\bm\theta_t}^\top
+
\bm J_{\bm x_t}
\bm\Sigma_{\Delta\bm x_t,\Delta\bm\theta}
\bm J_{\bm\theta_t}^\top
+
\bm J_{\bm\theta_t}
\bm\Sigma_{\Delta\bm\theta,\Delta\bm x_t}
\bm J_{\bm x_t}^\top.
\label{eq:Sigma_y_update_expanded}
\end{align}

The state--parameter cross-covariance must also be propagated from one step
to the next. Using the linearized perturbation dynamics, we obtain the
recursive update
\begin{equation}
\bm\Sigma_{\Delta\bm x_{t+1},\Delta\bm\theta}
=
\bm J_{\bm x_t}
\bm\Sigma_{\Delta\bm x_t,\Delta\bm\theta}
+
\bm J_{\bm\theta_t}
\bm\Sigma_{\Delta\bm\theta}.
\label{eq:cross_coF_update}
\end{equation}
Given an initial joint distribution $p(\bm x_0,\bm\theta)$, the recursive
updates \eqref{eq:mean_y_update}, \eqref{eq:Sigma_y_update}, and
\eqref{eq:cross_coF_update} can be used to propagate the approximate mean,
covariance, and state--parameter cross-covariance over multiple time steps.
At time $T$, the resulting moments $\bar{\bm x}_T$ and $\bm\Sigma_T$ define
a Gaussian approximation of the state PDF,
\begin{equation}
p_T(\bm x)
\approx
\mathcal{N}\left(\bar{\bm x}_T,\bm\Sigma_T\right).
\end{equation}
Since the network parameters $\bm \theta$ are fixed along each trajectory, 
their PDF, and hence their covariance $\bm\Sigma_{\Delta\bm\theta}$, 
remain unchanged during the forecast. Algorithm~\ref{alg:single_gaussian} 
summarizes this recursive procedure.

\begin{algorithm}[t]
\caption{Statistical Moments Propagation (Single Gaussian)}
\label{alg:single_gaussian}
\begin{algorithmic}[1]

\Require Initial joint distribution $p(\bm x_0,\bm\theta)$

\State Compute the means $\bar{\bm x}_0$ and $\bar{\bm\theta}$ and the
joint covariance
$\displaystyle
\bm\Sigma_{\Delta_0}
=
\operatorname{Cov}
\left(
\begin{bmatrix}
\Delta\bm x_0\\
\Delta\bm\theta
\end{bmatrix}
\right)
$
from $p(\bm x_0,\bm\theta)$

\For{$t=0,\ldots,T-1$}

    \State
    $\bar{\bm x}_{t+1}
    \leftarrow
    \bm F_{\bar{\bm\theta}}(\bar{\bm x}_t)$
    \Comment{Update state mean, Eq.~\eqref{eq:mean_y_update}}

    \State
    $\bm J_t
    \leftarrow
    [\bm J_{\bm x_t}\quad\bm J_{\bm\theta_t}]$
    \Comment{Compute combined Jacobian}

    \State
    $\bm\Sigma_{t+1}
    \leftarrow
    \bm J_t\bm\Sigma_{\Delta_t}\bm J_t^\top$
    \Comment{Update state covariance, Eq.~\eqref{eq:Sigma_y_update}}

    \State
    $\bm\Sigma_{\Delta\bm x_{t+1},\Delta\bm\theta}
    \leftarrow
    \bm J_{\bm x_t}
    \bm\Sigma_{\Delta\bm x_t,\Delta\bm\theta}
    +
    \bm J_{\bm\theta_t}
    \bm\Sigma_{\Delta\bm\theta}$
    \Comment{Update cross-covariance, Eq.~\eqref{eq:cross_coF_update}}

    \State Construct $\bm\Sigma_{\Delta_{t+1}}$ using
    $\bm\Sigma_{t+1}$,
    $\bm\Sigma_{\Delta\bm\theta}$, and
    $\bm\Sigma_{\Delta\bm x_{t+1},\Delta\bm\theta}$

\EndFor

\State \Return
$\{\bar{\bm x}_t,\bm\Sigma_t\}_{t=0}^{T}$ defining
$p(\bm x_t)\approx
\mathcal{N}(\bar{\bm x}_t,\bm\Sigma_t)$

\end{algorithmic}
\end{algorithm}

\subsection{Propagation Methods}
\label{sec:propagation_methods}
In this section, we present several methods for propagating the joint
distribution $p(\bm x_t,\bm\theta)$ forward in time through the
auto-regressive neural network. The goal is to approximate the marginal
state distribution $p(\bm x_t)$ at each time step.

We begin by representing the evolving state distribution by a collection of
$S$ samples, $\{\bm x_t^i\}_{i=1}^{S}$, at each time
$t=0,\ldots,T$. These samples form a point cloud whose mean and covariance
are estimated as
\begin{equation}
\bar{\bm x}_t
=
\frac{1}{S}\sum_{i=1}^{S}\bm x_t^i,
\qquad
\bm\Sigma_t
=
\frac{1}{S}\sum_{i=1}^{S}
\left(\bm x_t^i-\bar{\bm x}_t\right)
\left(\bm x_t^i-\bar{\bm x}_t\right)^\top.
\label{eq:mc_moments}
\end{equation}
Next, consider a standard Monte Carlo method in which
samples are propagated through the full, nonlinear neural network. At time
$t$, the state distribution is the push-forward of the initial joint
distribution $p(\bm x_0,\bm\theta)$ through the $t$-fold auto-regressive
composition $\bm F_{\bm\theta}^{(t)}=\underbrace{\bm F_{\bm\theta}\circ \cdots \circ \bm F_{\bm\theta}}_{\text{$t$ times}}$,
\begin{equation}
p(\bm x_t)
=
\int\int
\delta\left(
\bm x_t-\bm F_{\bm\theta}^{(t)}(\bm x_0)
\right)
p(\bm x_0,\bm\theta)
\,d\bm x_0\,d\bm\theta.
\label{eq:pushforward}
\end{equation}
This expression has the same form as the integral representation
\eqref{eq:PDF_g_general_dirac}. Drawing $S$ samples
$\{(\bm x_0^i,\bm\theta^i)\}_{i=1}^{S}$ from
$p(\bm x_0,\bm\theta)$ and propagating each sample auto-regressively for
$T$ steps yields the empirical distribution
\begin{equation}
p(\bm x_t)
\approx
\frac{1}{S}
\sum_{i=1}^{S}
\delta\left(\bm x_t-\bm x_t^i\right).
\label{eq:mc_pdf}
\end{equation}
The resulting Monte Carlo procedure is summarized in
Algorithm~\ref{alg:mc}.

\begin{algorithm}
\caption{Monte Carlo propagation with the full nonlinear network}
\label{alg:mc}
\begin{algorithmic}[1]

\Require Initial joint distribution $p(\bm x_0,\bm\theta)$
\Ensure State samples
$\{\bm x_t^i\}_{t=0,\ldots,T;\,i=1,\ldots,S}$

\State Sample
$\{(\bm x_0^i,\bm\theta^i)\}_{i=1}^{S}$
from $p(\bm x_0,\bm\theta)$

\For{$t=0,\ldots,T-1$}
    \For{$i=1,\ldots,S$}
        \State
        $\bm x_{t+1}^i
        \leftarrow
        \bm F_{\bm\theta^i}(\bm x_t^i)$
    \EndFor
\EndFor

\State \Return
$\{\bm x_t^i\}_{t=0,\ldots,T;\,i=1,\ldots,S}$

\end{algorithmic}
\end{algorithm}

The moment-propagation equations developed in
Section~\ref{sec:Auto-regressive_Statistical_Moments} characterize the
evolving distribution only through its first two moments. Consequently,
when the initial distribution is non-Gaussian, representing the propagated
state distribution by a single Gaussian introduces an additional
approximation beyond the linearization of the neural network. Moreover, in
our numerical experiments, the single-Gaussian propagation of
Algorithm~\ref{alg:single_gaussian} produces variances that grow rapidly over
relatively short time horizons, eventually leading to unstable predictions.

\subsubsection{Moment Propagation with Resampling}
\label{sec:rmp}

Rather than representing the evolving state distribution by a single
Gaussian, we represent $p(\bm x_t)$ by an empirical distribution of $S$
particles and alternate between two phases. During the propagation phase,
each particle is associated with a local Gaussian whose mean and covariance
are evolved using the linearized moment-propagation equations. During the
resampling phase, each particle is displaced by a sample drawn from its local
Gaussian, after which its covariance is reset to zero. In this way, the
evolving distribution is represented by a point cloud rather than a single
Gaussian, allowing non-Gaussian features to be retained while limiting the
covariance growth observed in the single-Gaussian propagation.

To describe the moment-propagation method with resampling, we initialize $S$
particles $\{\bm x_0^i\}_{i=1}^S$ by sampling from the marginal distribution
$p(\bm x_0)$. Each particle is initially treated as a point mass, so that
\begin{equation}
\bm\Sigma_0^i=\bm 0,
\qquad
\bm\Sigma_{\Delta\bm x_0,\Delta\bm\theta}^i=\bm 0,
\qquad i=1,\ldots,S.
\end{equation}
At each time step, the particles are propagated independently using the
moment-propagation equations derived in
Section~\ref{sec:Auto-regressive_Statistical_Moments}. For particle $i$, the
mean and covariance are updated according to
\begin{equation}
\bar{\bm x}_{t+1}^i
=
\bm F_{\bar{\bm\theta}}(\bm x_t^i),
\qquad
\bm\Sigma_{t+1}^i
=
\bm J_t^i
\bm\Sigma_{\Delta_t}^i
\bm J_t^{i\top},
\end{equation}
together with the corresponding state--parameter cross-covariance update
$\bm\Sigma_{\Delta\bm x_{t+1},\Delta\bm\theta}^i$ from
\eqref{eq:cross_coF_update}. Between resampling steps, 
the particle location is identified with the
propagated mean, $\bm x_{t+1}^i=\bar{\bm x}_{t+1}^i$, while its local
covariance accumulates the uncertainty induced by the network parameters
over successive time steps.

At a resampling step, the local Gaussian distributions associated with the
particles are used to generate a new point cloud, after which the local
covariances and state--parameter cross-covariances are reset. We consider two
resampling strategies. The first draws one sample from the local Gaussian
associated with each particle, thereby preserving the number of particles
$S$. The second treats the collection of local Gaussians as a Gaussian
mixture model and draws a prescribed number $L$ of samples from the resulting
mixture. We describe both strategies below.

\paragraph{One Sample per Gaussian}

At each resampling event, the current state distribution is represented by
the empirical approximation
\begin{equation}
\widehat{p}(\bm x_t)
=
\frac{1}{S}
\sum_{i=1}^{S}
\delta\left(\bm x_t-\bm x_t^i\right).
\end{equation}
Propagating each point mass using the moment equations derived in
Section~\ref{sec:Auto-regressive_Statistical_Moments} yields a local Gaussian
distribution,
\begin{equation}
p(\bm x_{t+1}\mid\bm x_t^i)
\approx
\mathcal{N}\left(
\bar{\bm x}_{t+1}^i,
\bm\Sigma_{t+1}^i
\right),
\label{eq:local_transition}
\end{equation}
where $\bar{\bm x}_{t+1}^i$ and $\bm\Sigma_{t+1}^i$ are obtained from the
moment-propagation equations. The resulting state distribution is therefore
approximated by the Gaussian mixture
\begin{equation}
p(\bm x_{t+1})
\approx
\frac{1}{S}
\sum_{i=1}^{S}
\mathcal{N}\left(
\bar{\bm x}_{t+1}^i,
\bm\Sigma_{t+1}^i
\right).
\end{equation}
The one-sample-per-Gaussian strategy replaces this mixture by a new point
cloud by drawing one sample from each component,
\begin{equation}
\bm x_{t+1}^i
\sim
\mathcal{N}\left(
\bar{\bm x}_{t+1}^i,
\bm\Sigma_{t+1}^i
\right),
\qquad i=1,\ldots,S.
\end{equation}
After resampling, the local state covariances and state--parameter
cross-covariances are reset to zero, and moment propagation resumes from the
new set of point masses.

\paragraph{Sampling from the Gaussian Mixture Model (GMM)}

At each resampling step, the collection of local Gaussian distributions
defines the equally weighted Gaussian mixture
\begin{equation}
p(\bm x_{t+1})
\approx
\frac{1}{S}
\sum_{i=1}^{S}
\mathcal{N}\left(
\bar{\bm x}_{t+1}^i,
\bm\Sigma_{t+1}^i
\right).
\end{equation}
To sample from this mixture, we first draw $L$ samples from each local
Gaussian,
\begin{equation}
\tilde{\bm x}_{t+1}^{i,k}
\sim
\mathcal{N}\left(
\bar{\bm x}_{t+1}^i,
\bm\Sigma_{t+1}^i
\right),
\qquad
k=1,\ldots,L,
\qquad
i=1,\ldots,S.
\end{equation}
This produces a combined pool of $SL$ samples. We then uniformly select $S$
samples from this pool to form the new point cloud
$\{\bm x_{t+1}^i\}_{i=1}^{S}$. Since each Gaussian component contributes the
same number $L$ of samples, uniform sampling from the combined pool
approximates sampling from the equally weighted Gaussian mixture.
After resampling, the new particles are treated as point masses: their local
state covariances and state--parameter cross-covariances are reset to zero,
and moment propagation resumes independently from each particle. This
propagation and resampling procedure is repeated until the final time
$T$ is reached.
For both resampling strategies, the state distribution at each time step is
represented by the empirical distribution
\begin{equation}
p(\bm x_t)
\approx
\frac{1}{S}
\sum_{i=1}^{S}
\delta\left(
\bm x_t-\bm x_t^i
\right).
\label{eq:empirical}
\end{equation}
The complete procedure is summarized in Algorithm~\ref{alg:rmp}.

\paragraph{Resampling Strategy}
The resampling interval is the primary parameter controlling the performance
of the method and determines a tradeoff between two failure modes. If the
interval is too large, the local covariances $\bm\Sigma_t^i$ can grow beyond
the regime in which the linearized model provides an accurate approximation,
eventually leading to instability. If the interval is too small, the local
covariances are reset before sufficient uncertainty can accumulate, causing
the particle cloud to remain overly concentrated and systematically
underestimate uncertainty.
At each resampling step, we draw exactly one sample from each Gaussian
component. This guarantees that every component contributes one descendant,
so no particle lineage is lost during resampling. Alternatively, the local
Gaussians can be treated as an equally weighted Gaussian mixture by drawing
$L>1$ samples from each component and uniformly subsampling the resulting
pool back to $S$ particles. In this case, some components may contribute
multiple descendants while others may contribute none.
We observed empirically that combining frequent resampling with $L>1$ can
cause the particle cloud to progressively collapse over repeated resampling
steps, resulting in sample impoverishment or particle degeneracy
(see ~\ref{app:particle_degeneration}). When the resampling interval
is sufficiently large, however, the local covariances $\bm\Sigma_t^i$
accumulate enough spread that the $L>1$ strategy performs comparably to the
one-sample-per-Gaussian strategy.
The distinction between enforcing one descendant from each component and
sampling freely from the mixture is closely related to the distinction
between stratified and multinomial resampling in particle filtering. In both
cases, stratification limits the loss of representation caused by random
resampling. Consistent with this interpretation, stratified resampling has
been shown to reduce estimator variance relative to multinomial resampling,
including in the equal-weight setting~\cite{douc_comparison_2005}.

\begin{algorithm}[t]
\caption{Resampled Moment Propagation}
\label{alg:rmp}
\begin{algorithmic}[1]

\Require Initial joint distribution $p(\bm x_0,\bm\theta)$,
resampling interval $\tau$, and number of local samples $L$

\Ensure State samples
$\{\bm x_t^i\}_{t=0,\ldots,T;\,i=1,\ldots,S}$

\State Sample $\{\bm x_0^i\}_{i=1}^{S}$ from the marginal $p(\bm x_0)$
\State Set
$\bm\Sigma_0^i=\bm 0$ and
$\bm\Sigma_{\Delta\bm x_0,\Delta\bm\theta}^i=\bm 0$
for all $i$

\For{$t=0,\ldots,T-1$}

    \For{$i=1,\ldots,S$}

        \State
        $\bar{\bm x}_{t+1}^i
        \leftarrow
        \bm F_{\bar{\bm\theta}}(\bm x_t^i)$
        \Comment{Update mean}

        \State
        $\bm J_t^i
        \leftarrow
        [\bm J_{\bm x_t^i}\quad\bm J_{\bm\theta_t^i}]$
        \Comment{Compute combined Jacobian}

        \State
        $\bm\Sigma_{t+1}^i
        \leftarrow
        \bm J_t^i
        \bm\Sigma_{\Delta_t}^i
        {\bm J_t^i}^{\top}$
        \Comment{Update state covariance}

        \State
        $\bm\Sigma_{\Delta\bm x_{t+1},\Delta\bm\theta}^i
        \leftarrow
        \bm J_{\bm x_t^i}
        \bm\Sigma_{\Delta\bm x_t,\Delta\bm\theta}^i
        +
        \bm J_{\bm\theta_t^i}
        \bm\Sigma_{\Delta\bm\theta}$
        \Comment{Update cross-covariance}

    \EndFor

    \If{$(t+1)\bmod\tau=0$}

        \For{$i=1,\ldots,S$}
            \State
            $\{\tilde{\bm x}_{t+1}^{i,k}\}_{k=1}^{L}
            \sim
            \mathcal{N}\!\left(
            \bar{\bm x}_{t+1}^i,
            \bm\Sigma_{t+1}^i
            \right)$
            \Comment{Generate local samples}
        \EndFor

        \If{$L=1$}
            \State
            $\bm x_{t+1}^i
            \leftarrow
            \tilde{\bm x}_{t+1}^{i,1}$
            for all $i$
            \Comment{One sample per Gaussian}
        \Else
            \State Uniformly select $S$ samples from
            $\displaystyle
            \bigcup_{i=1}^{S}
            \{\tilde{\bm x}_{t+1}^{i,k}\}_{k=1}^{L}$
            to form $\{\bm x_{t+1}^i\}_{i=1}^{S}$
            \Comment{Sample from GMM}
        \EndIf

        \State Set
        $\bm\Sigma_{t+1}^i=\bm 0$ and
        $\bm\Sigma_{\Delta\bm x_{t+1},\Delta\bm\theta}^i=\bm 0$
        for all $i$
        \Comment{Reset to point masses}

    \Else

        \State
        $\bm x_{t+1}^i
        \leftarrow
        \bar{\bm x}_{t+1}^i$
        for all $i$

    \EndIf

\EndFor

\State \Return
$\{\bm x_t^i\}_{t=0,\ldots,T;\,i=1,\ldots,S}$

\end{algorithmic}
\end{algorithm}

\section{Numerical Experiments}
\label{sec:numerical_experiments}

In this section, we apply the auto-regressive propagation methods discussed
in Section~\ref{sec:propagation_methods} to MLPs modeling autonomous chaotic
dynamical systems. The numerical experiments examine how these methods
represent the evolution of uncertainty in both the initial state and network
parameters over successive (auto-regressive) applications of the MLP.
Specifically, we consider the Lorenz-63 \cite{lorenz1963deterministic} 
and the Kuramoto--Sivashinsky \cite{temam1997infinite,nicolaenko1985global} systems. 
For appropriate choices of their parameters, both systems exhibit chaotic 
dynamics characterized by strange attractors with positive Lyapunov 
exponents, leading to a strong sensitivity to initial conditions.
The MLPs representing the one-step dynamics have Leaky ReLU activation
functions and are trained on unperturbed solutions of the Lorenz-63 and Kuramoto--Sivashinsky
differential equations. The solutions are generated using RK45 with adaptive
time stepping and then down-sampled to the temporal discretizations used for
training. The network architectures and training parameters are summarized
in Table~\ref{tab:nn_setup}.

The parameters defining the initial Gaussian distributions for each system
are given in Table~\ref{tab:initial_distributions_scaling}. The state
perturbations have zero mean, with magnitudes chosen to produce an 
appreciable growth of uncertainty over relatively short time horizons.
For the network parameters, we also consider zero-mean perturbations, with
standard deviations estimated from parameter fluctuations during the late
stages of training. These standard deviations are then scaled to ensure that
the contribution of parameter uncertainty is not negligible relative to
initial-state uncertainty. The initial-state and parameter perturbations are
assumed to be statistically independent at the initial time. Under
auto-regressive propagation, however, the state and network parameters become
statistically dependent after a single time step.

\begin{table}[t]
\centering
\begin{tabular}{lll}
\hline
\textbf{Component}          & \textbf{Lorenz-63}           &  \textbf{Kuramoto--Sivashinsky} \\
\hline
Dimension of system         & 3                         & 200 \\
Number of hidden layers     & 3                         & 4   \\
Hidden width                & 64 neurons per layer      & 256 neurons per layer\\
Total parameters       & $\approx$ 9,000              & $\approx$ 300,000 \\
Batch size                  & 1024                      & 256 \\
Number of training samples  & 100,000                   & 500,000\\
Epochs                      & 1000                      & 500 \\
%Learning rate scheduler     & \texttt{ReduceLROnPlateau} & \\
\hline
\end{tabular}
\caption{Neural network training setup for the Lorenz-63 and Kuramoto--Sivashinsky systems.
Both models use the MLP architecture described in 
Section~\ref{sec:background} and are trained using the \texttt{Adam} optimizer with the
\texttt{torch.nn.MSELoss} loss function. The data are split into 70\% training,
20\% validation, and 10\% testing sets. The regression data modules are based
on those provided in \texttt{lightning-uq-box}~\cite{lightninguq_github}.}
\label{tab:nn_setup}
\end{table}

\begin{table}[t]
\centering
\begin{tabular}{lcc}
\hline
\textbf{} & \textbf{Lorenz-63} & \textbf{Kuramoto--Sivashinsky} \\
\hline
\multicolumn{3}{l}{{\em Initial state}} \\
\hline
Standard deviation & $10^{-3}$ & $10^{-1}$ \\
\hline
\multicolumn{3}{l}{{\em MLP  parameters }} \\
\hline
Epochs used & 30 (3\% of total) & 30 (6\% of total) \\
Base standard dev. & $\sim 1.28 \times 10^{-6}$ & $\sim 2.94\times 10^{-5}$ \\
Scaling factor & 100 & 5 \\
Effective standard dev. & $1.28 \times10^{-4}$ & $1.47\times 10^{-4}$ \\
\hline
\end{tabular}
\caption{Standard deviations of the initial-state and network-parameter
perturbations used in the Lorenz-63 and Kuramoto--Sivashinsky uncertainty-propagation
experiments. The perturbation magnitudes are chosen to produce appreciable
uncertainty growth over relatively short time horizons while ensuring a
non-negligible contribution from parameter uncertainty relative to
initial-state uncertainty. The initial-state and parameter perturbations are
assumed to be statistically independent. Under auto-regressive propagation,
however, the state and network parameters become statistically dependent
after a single propagation step.}
\label{tab:initial_distributions_scaling}
\end{table}

\subsection{Lorenz-63 System}
\label{sec:lorenz}

Consider the Lorenz-63 system
\begin{equation}
\begin{cases}
    \dfrac{dX}{dt} = \sigma(Y-X), \\[8pt]
    \dfrac{dY}{dt} = X(\rho-Z)-Y, \\[8pt]
    \dfrac{dZ}{dt} = XY-\beta Z,
\end{cases}
\label{eq:lorenz}
\end{equation}
with parameters $\sigma=10$, $\beta=8/3$, and $\rho=28$. For this parameter
regime, the system exhibits chaotic dynamics characterized by bounded,
aperiodic trajectories and strong sensitivity to initial conditions.
The largest Lyapunov exponent is approximately $\ell_1=0.902$, corresponding
to a Lyapunov time of
\begin{equation}
T_L=\frac{1}{\ell_1}\approx 1.11.
\end{equation}
Using a one-hundred-fold amplification of an initial perturbation as the
predictability criterion gives the predictability horizon
\begin{equation}
T_p=\frac{\log(100)}{\ell_1}\approx 5.11.
\end{equation}
We solve \eqref{eq:lorenz} using RK45 with adaptive time stepping and
down-sample the resulting solution to a temporal resolution of
$\Delta t=0.01$.

\begin{figure}[t]
    \centering

    \begin{minipage}[t]{.55\linewidth}
    \centering
    \subcaption{(a)}\label{fig:timeseries_lorenz_mc}
    \end{minipage}%
    \begin{minipage}[t]{.45\linewidth}
    \centering
    \subcaption{(b)}\label{fig:timeseries_lorenz_part}
    \end{minipage} 

    \includegraphics[width=1\linewidth]{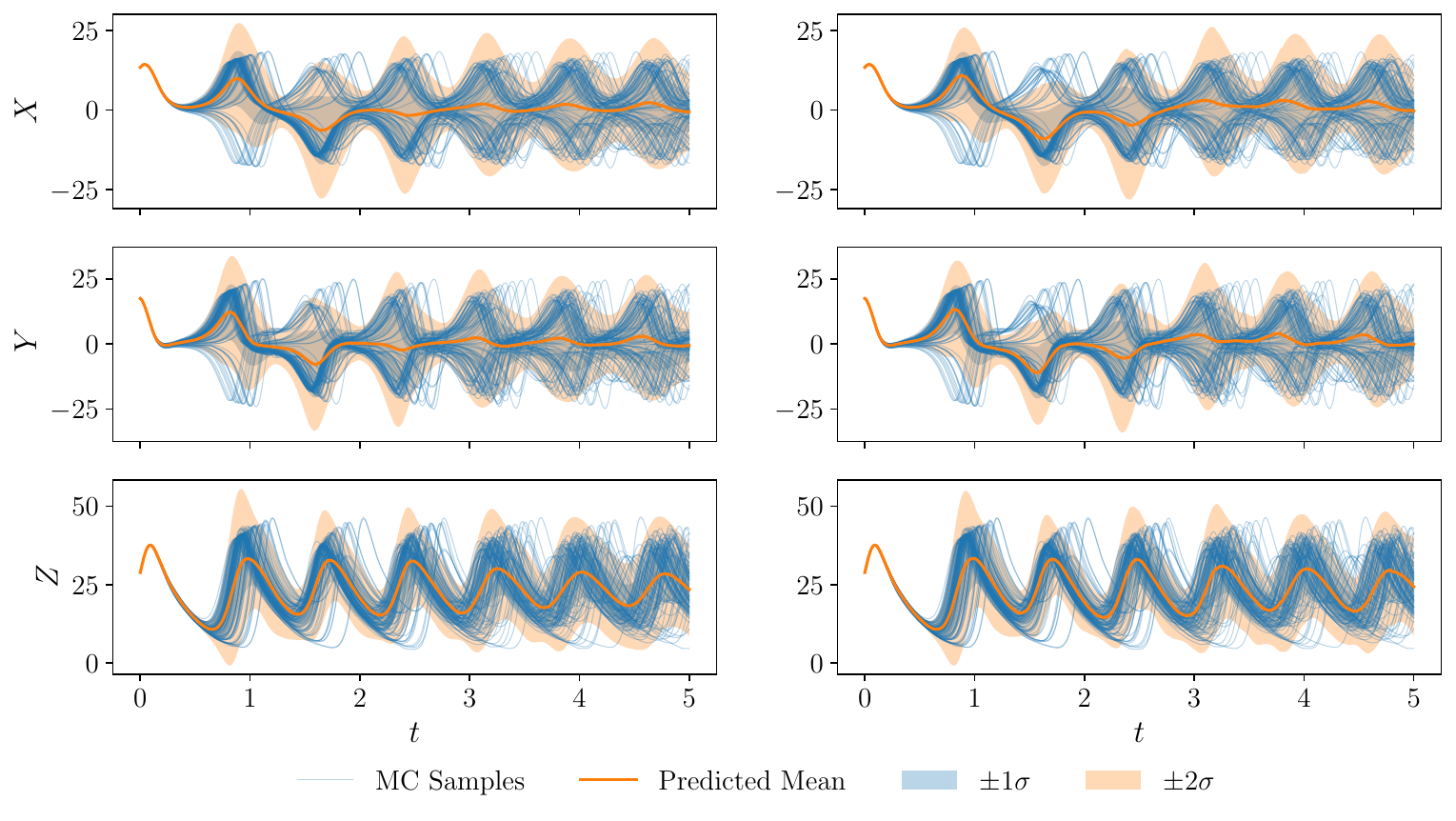}

\caption{Uncertainty propagation over time for the Lorenz-63 system. The blue
trajectories in both panels are samples from the Monte Carlo baseline,
obtained by propagating the initial samples through the full nonlinear
network. Panel (a) shows the mean and standard deviation computed from the
Monte Carlo ensemble, while panel (b) shows the corresponding statistics
obtained using Resampled Moment Propagation. Despite relying on a
local approximation of the network, Resampled Moment Propagation closely
reproduces the Monte Carlo mean and covariance over the full forecast horizon.
The particle method uses $S=1000$ particles and a resampling interval of 20
time steps. At each resampling step, $L=100$ samples are drawn from the local
Gaussian associated with each particle, and the resulting sample pool is
uniformly subsampled back to $S=1000$ particles.}
\label{fig:lorenz_timeseries_main}
\end{figure}

\begin{figure}[t]
    \centering

    \begin{minipage}[t]{.55\linewidth}
    \centering
    \subcaption{(a)}\label{fig:joint_lorenz_mc}
    \end{minipage}%
    \begin{minipage}[t]{.45\linewidth}
    \centering
    \subcaption{(b)}\label{fig:joint_lorenz_part}
    \end{minipage} 

    \includegraphics[width=1\linewidth]{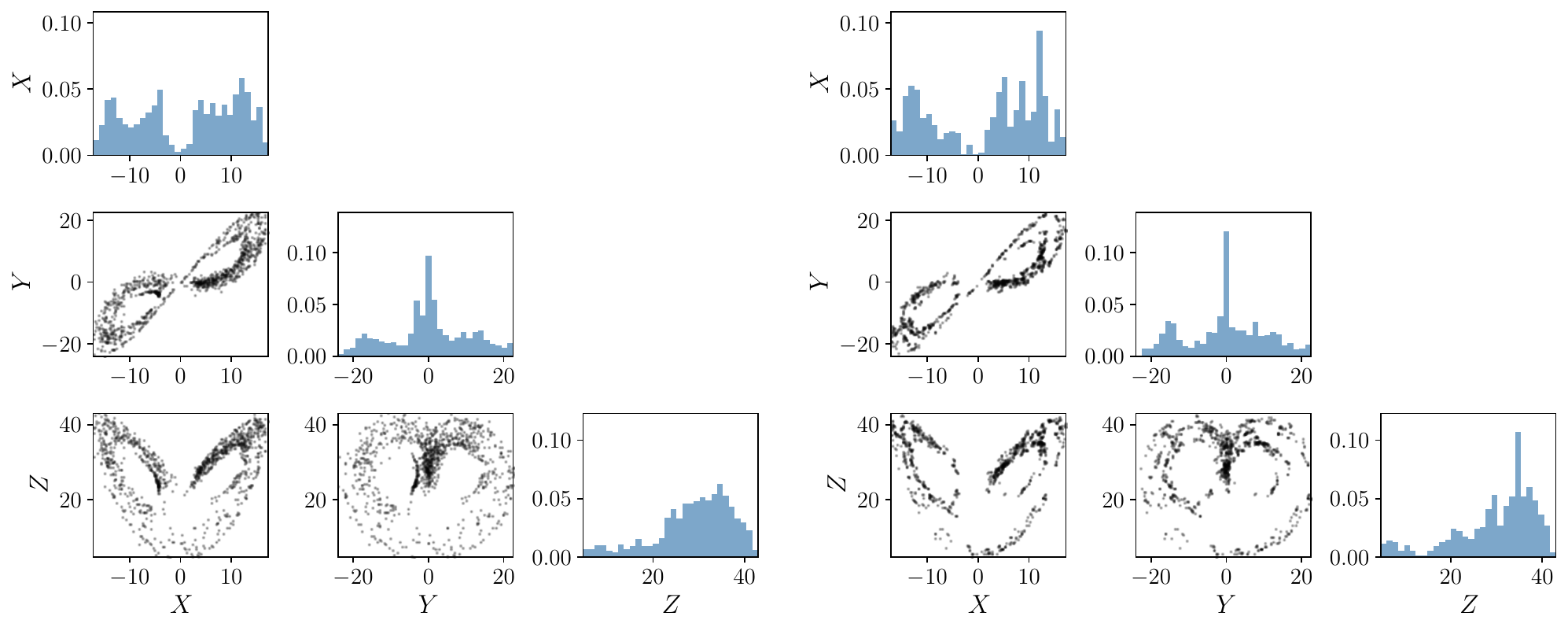}

\caption{Joint and marginal distributions for the Lorenz-63 system at
$t=4$, corresponding to 400 time steps. The particle clouds correspond to
(a) the Monte Carlo baseline and (b) Resampled Moment Propagation. The
Resampled Moment Propagation configuration is the same as in
Figure~\ref{fig:lorenz_timeseries_main}, with $S=1000$ particles, $L=100$
local samples drawn from each particle at each resampling step, and a
resampling interval of 20 time steps.}
    \label{fig:lorenz_joint_main}
\end{figure}

\begin{figure}[t]
    \centering

    \begin{minipage}[t]{.55\linewidth}
    \centering
    \subcaption{(a)}\label{fig:timeseries_lorenz_mc100}
    \end{minipage}%
    \begin{minipage}[t]{.45\linewidth}
    \centering
    \subcaption{(b)}\label{fig:timeseries_lorenz_part100}
    \end{minipage} 

    \includegraphics[width=1\linewidth]{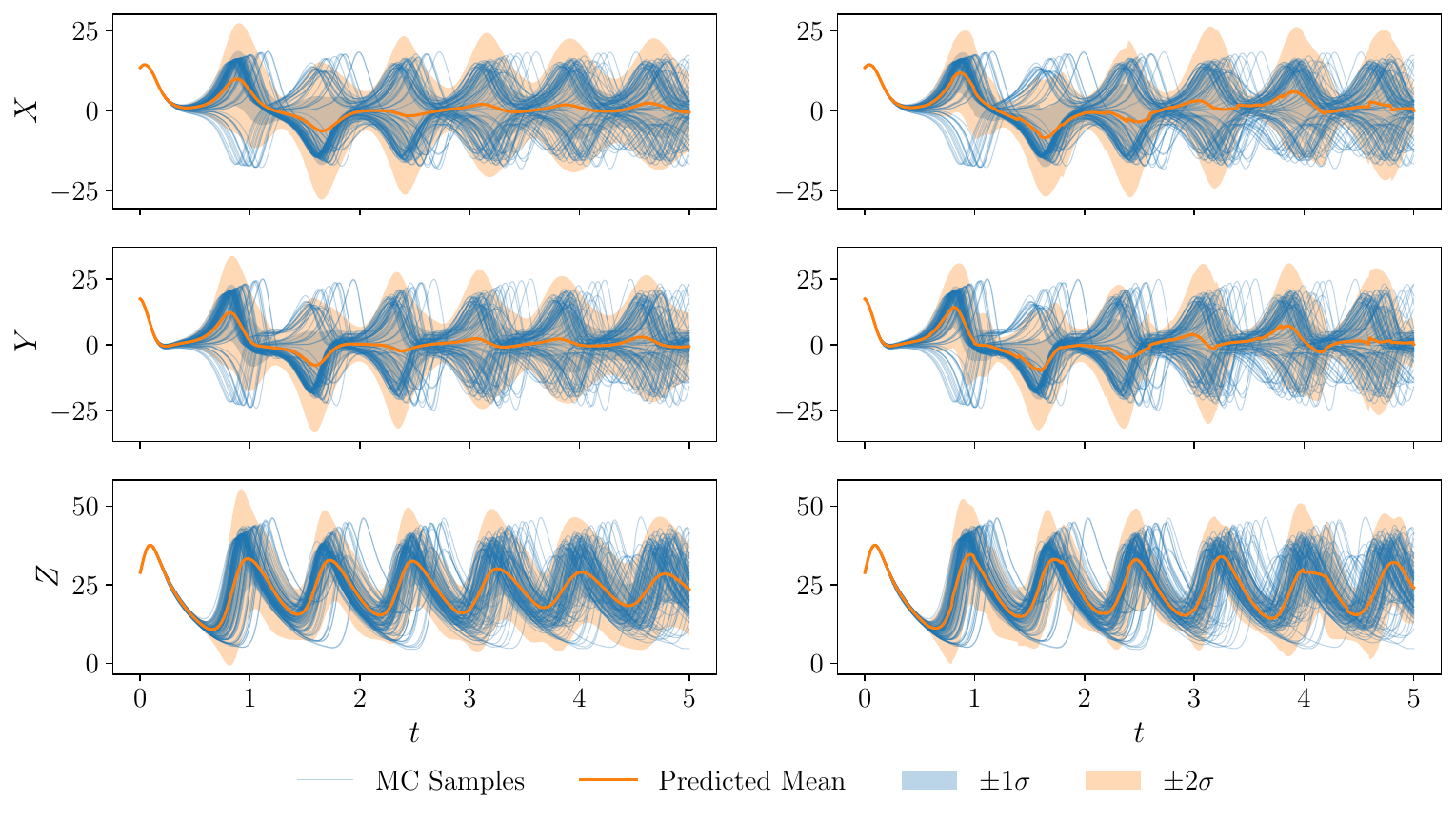}

 \caption{Uncertainty propagation over time for the Lorenz-63 system. This is the same as Figure \ref{fig:lorenz_timeseries_main}, but with $S=100$ particles. The local samples and resampling steps are the same ($L=100$, and resampling every $20$ steps).}
    
    \label{fig:lorenz_timeseries_S100}
\end{figure}

The single-Gaussian propagation of Algorithm~\ref{alg:single_gaussian}
initially overestimates the variance and then becomes unstable around
$t\approx 2$, with the predicted variance growing rapidly and far exceeding
the spread of the Monte Carlo ensemble. Representing the distribution by a
Gaussian mixture with a small number of components (10--20), without
resampling, does not significantly improve the results.
In contrast, the Resampled Moment Propagation method presented in
Section~\ref{sec:rmp} mitigates the growth of the predicted variance through
periodic resampling, whereby the accumulated local covariances are reset to
zero. As discussed in Section~\ref{sec:rmp}, resampling at every step without
forcing one sample per component may lead to severe sample impoverishment,
causing the particle cloud to collapse and the variance to be underestimated.
When the resampling interval is sufficiently large to allow appreciable
covariance accumulation, the one-sample-per-component and GMM sampling
strategies produce comparable results. This indicates that the resampling
interval is indeed the primary parameter controlling the performance 
of the method.
For the Lorenz-63 system, we find that resampling intervals ranging from
approximately 10 to 50 time steps produce comparable results. Within this
range, the number of local samples $L$ has little effect. For shorter
resampling intervals, however, particle degeneracy becomes more pronounced
as $L$ increases, approaching the limiting behavior discussed in
Section~\ref{sec:rmp}.

Figures~\ref{fig:lorenz_timeseries_main} and~\ref{fig:lorenz_joint_main}
show the trajectory rollouts obtained using $S=1000$ particles with the
Resampled Moment Propagation method, compared with $3000$ samples for the
Monte Carlo baseline. We note that, in our experiments, 
as few as $100$ particles produced comparable results for resampling 
intervals ranging from approximately 10--50 steps (see Figure~\ref{fig:lorenz_timeseries_S100}).
Figure~\ref{fig:lorenz_joint_main} shows the joint and marginal distributions
of the Lorenz-63 system at $t=4$ (400 time steps), comparing the particle
clouds obtained with the Monte Carlo baseline (a), and our Resampled Moment
Propagation method (b). It is seen that despite relying on local approximation, 
Resampled Moment Propagation closely reproduces the mean, covariance,
and overall distribution obtained by Monte Carlo, and therefore provides accurate
uncertainty estimates through the predictability horizon $T_p\approx5.11$.
For a naive implementation, a full rollout of the Resampled Moment Propagation method over 500 steps with 100 particles takes approximately 2 minutes, scaling to roughly 20 minutes for 1000 particles. Runtime scales linearly with the number of particles $S$ and is independent of the resampling interval.

\subsection{Kuramoto--Sivashinsky System}
\label{sec:KS}

The Kuramoto--Sivashinsky (KS) equation is a fourth-order nonlinear PDE that provides a
canonical model of spatiotemporal chaotic dynamics
\cite{temam1997infinite,nicolaenko1985global}. In particular, we consider the
initial-boundary-value problem
\begin{equation}
    \begin{cases}
        \dfrac{\partial u}{\partial t}
        + u\dfrac{\partial u}{\partial y}
        + \dfrac{\partial^2 u}{\partial y^2}
        + \dfrac{\partial^4 u}{\partial y^4}
        = 0,
        & t \geq 0, \quad y \in [-25,25], \\[10pt]
        u(y,0) = \sin(y)\,e^{-(y-10)^2/2}, \\[8pt]
        \textrm{Periodic boundary conditions}.
    \end{cases}
    \label{eq:KS}
\end{equation}
Unlike the Lorenz-63 system, the KS equation defines an infinite-dimensional
dynamical system. To obtain a finite-dimensional representation, we
discretize the spatial domain using $d=200$ equally spaced grid points on
$[-25,25]$, resulting in a 200-dimensional system of coupled ODEs. This system is
integrated using RK45 with adaptive time stepping, and the resulting solution
is sampled at a temporal resolution of $\Delta t=0.02$.
The largest Lyapunov exponent for this discretization is approximately $\ell_1 \approx 
0.11$, giving a Lyapunov time of $T_L = 1 /\ell_1 \approx 9.09$. For a one-hundred-fold amplification, this gives a predictability horizon of $T_p = \log(100) / \ell_1 \approx 41.84$.

\begin{figure}[ht]
    \centering
    \begin{minipage}[t]{.55\linewidth}
    \centering
    \subcaption{(a)}\label{fig:timeseries_ks}
    \end{minipage}%
    \begin{minipage}[t]{.45\linewidth}
    \centering
    \subcaption{(b)}\label{fig:timeseries_ks_part}
    \end{minipage} 

    \includegraphics[width=1\linewidth]{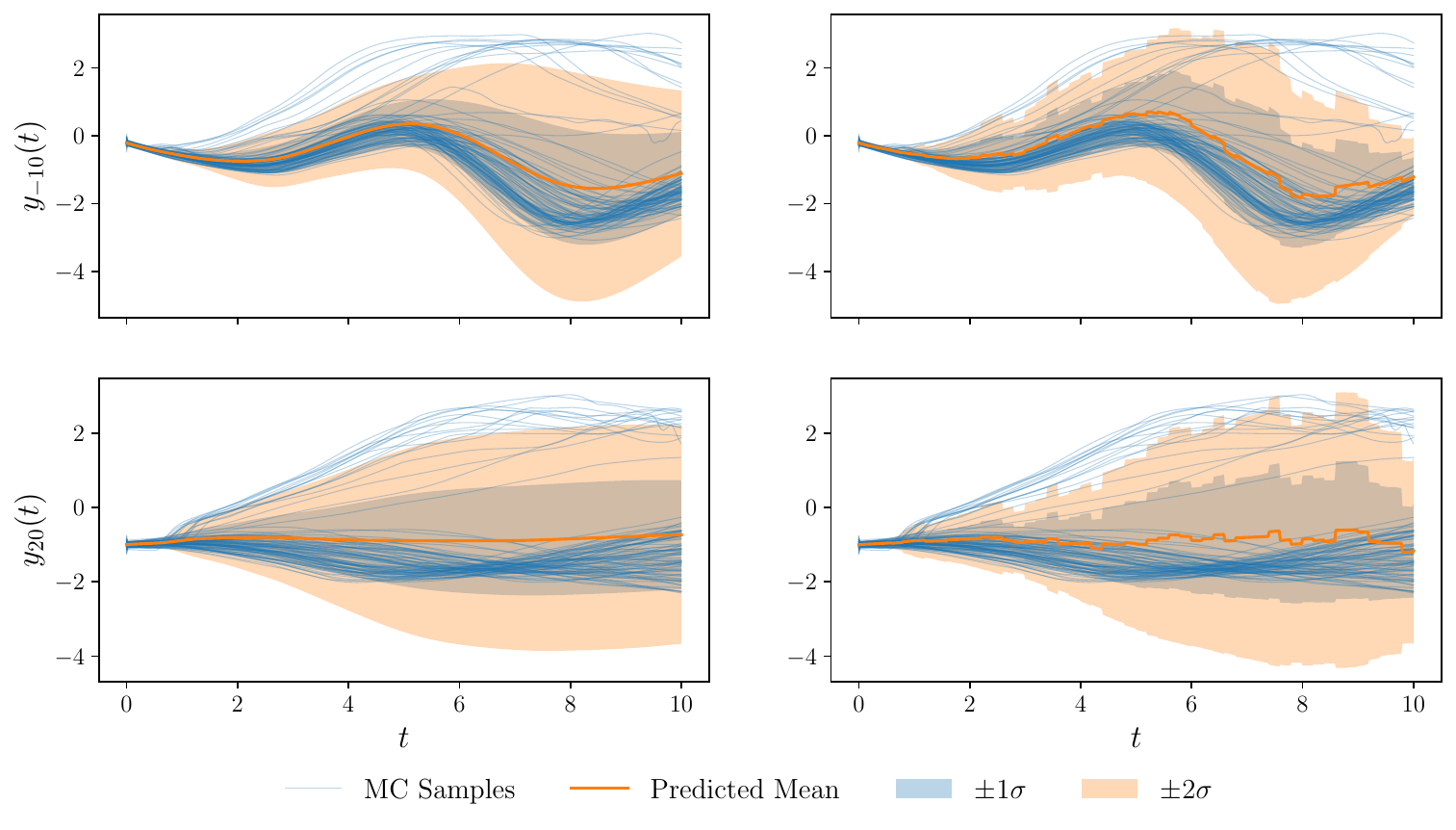}

 \caption{Uncertainty propagation over time for the Kuramoto--Sivashinsky system. The blue
trajectories in both panels are samples from the Monte Carlo baseline,
obtained by propagating the initial samples through the full nonlinear
network. Panel (a) shows the mean and standard deviation computed from the
Monte Carlo ensemble, while panel (b) shows the corresponding statistics
obtained using Resampled Moment Propagation. The particle method uses
$S=100$ particles and a resampling interval of 10 time steps. At each
resampling step, $L=100$ samples are drawn from the local Gaussian associated
with each particle, and the resulting sample pool is uniformly subsampled
back to $S=100$ particles.}
    
    \label{fig:ks_timeseries_main}
\end{figure}

\begin{figure}[ht]
    \centering
    \centerline{\footnotesize \hspace{0.4cm}(a)\hspace{4.2cm}(b)\hspace{5.2cm} (c)}

\centerline{
\rotatebox{90}{\hspace{1.6cm}\footnotesize Standard Deviation \hspace{2.9cm} Mean }
    \includegraphics[width=1\linewidth]{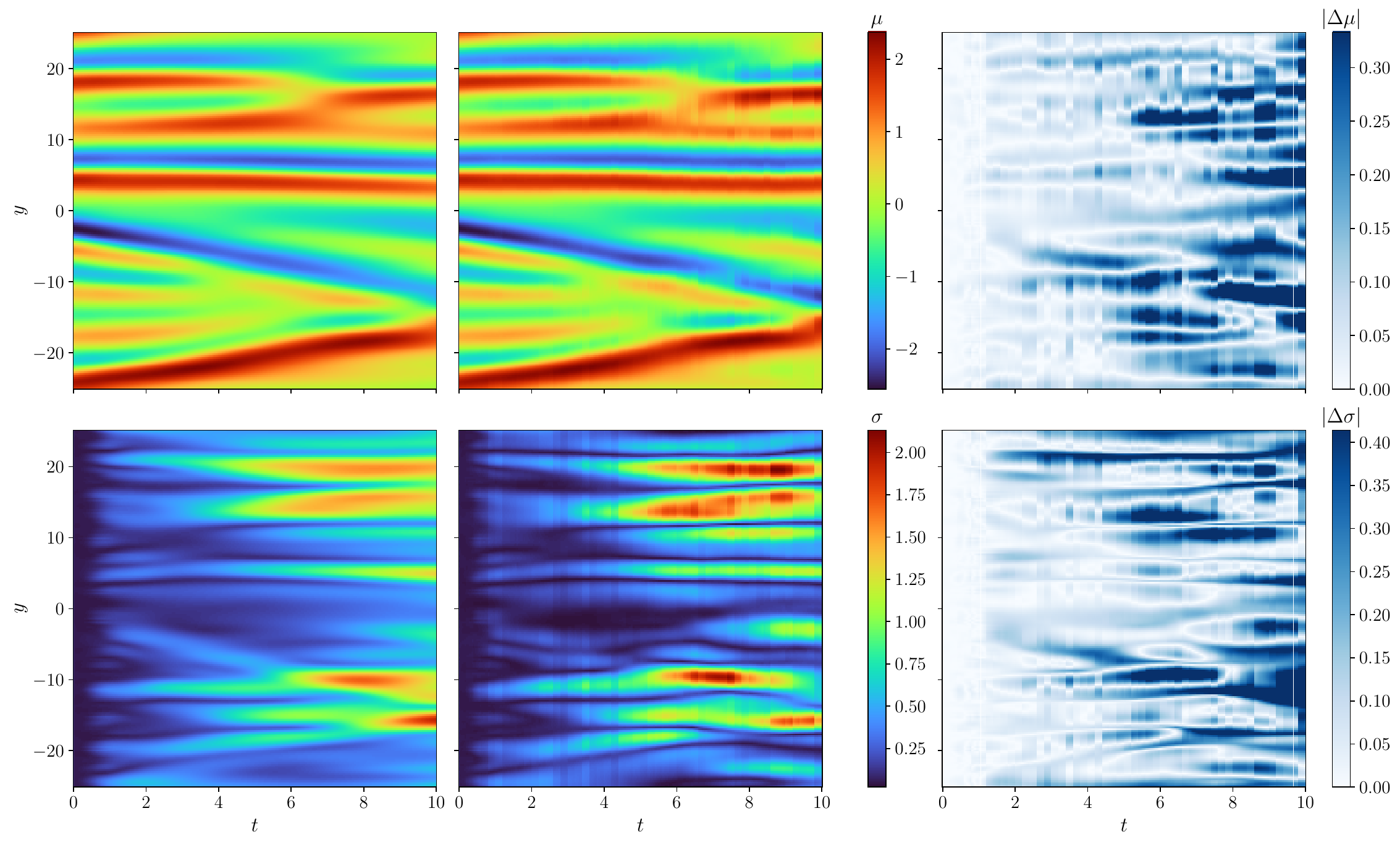}}

\caption{Mean (top) and one standard deviation (bottom) for the Kuramoto--Sivashinsky system
over 10 seconds, corresponding to 500 time steps.  Columns (a) and (b) show
the statistics obtained from the Monte Carlo baseline and Resampled Moment
Propagation, respectively, while column (c) shows their absolute difference.
The Resampled Moment Propagation method uses $S=100$ 
particles and a resampling interval of 10 time steps. At each 
resampling step, $L=100$ samples are drawn from the local
Gaussian associated with each particle, and the resulting sample pool is
uniformly subsampled back to $S=100$ particles.}

    \label{fig:ks_heatmap}
\end{figure}

For the Resampled Moment Propagation experiment, we use a resampling interval
of 10 time steps and $S=100$ particles, compared with $S=3000$ samples for
the Monte Carlo baseline. The rollout spans 500 time steps, corresponding to
only a fraction of the predictability horizon of the KS system. To obtain an
appreciable growth of uncertainty within this shorter time window, we increase
the scaling of the initial parameter variance, as reported in
Table~\ref{tab:initial_distributions_scaling}. We emphasize that $S=100$
particles provides a very sparse representation of the 200-dimensional state
space.
 
The primary practical limitation of the method at this scale is memory.
The state--parameter cross-covariance
$\bm{\Sigma}^i_{\Delta\bm{x},\Delta\bm{\theta}}$ has dimensions
$(S,d,N_{\theta})=(100,200,300{,}000)$ and requires approximately
48\,GB of storage in double precision. This represents the storage required
for the cross-covariances alone and does not include additional quantities,
such as the particle-specific Jacobians.
We tested a reduced variant in which the state--parameter cross-covariance
is neglected entirely. In this case, each particle requires only a
$200\times200$ state covariance matrix and a $300{,}000$-dimensional
parameter variance vector. This approximation, however, substantially
underestimates the propagated variance. Similarly, as discussed in
Sections~\ref{sec:rmp} and~\ref{sec:lorenz}, setting the resampling 
period to 1 eliminates the need to store and propagate the
cross-covariance, since it is reset at every step, but also leads to a
substantial underestimation of the variance.
With a naive implementation, the full 500-step rollout with 100 particles required approximately
13 hours, compared with roughly 2 minutes for the Lorenz-63 system over the
same number of time steps. Between resampling steps, however, the $S$
particles evolve independently and can therefore be propagated in parallel,
providing a straightforward means of substantially reducing the wall-clock
time.
Despite these computational constraints, the method accurately tracks the
mean and captures the initial growth of the variance. Small discontinuities
in the predicted variance are visible at the resampling steps, as expected
from resetting the local covariances to zero.
Figures~\ref{fig:ks_timeseries_main} and~\ref{fig:ks_heatmap} compare the
uncertainty propagation obtained with the Monte Carlo baseline and Resampled
Moment Propagation for the Kuramoto--Sivashinsky system. Figure~\ref{fig:ks_timeseries_main}
shows the evolution of the statistics, with (a) corresponding to the Monte
Carlo ensemble and (b) to Resampled Moment Propagation.
Figure~\ref{fig:ks_heatmap} shows the mean (top) and one standard deviation
(bottom) over 10 seconds (500 time steps), with (a) the Monte Carlo baseline,
(b) Resampled Moment Propagation, and (c) the absolute difference between the
two.

\section{Conclusion}
\label{sec:conclusion}

We developed a new framework for uncertainty propagation 
in random MLPs in which both the input and network parameters 
are random. Building upon the piecewise-linear structure of the 
Leaky ReLU activation function, we derived a local approximation 
of the neural network with respect to perturbations in both 
its inputs and parameters. This representation is
exact with respect to input perturbations that preserve the network activation
pattern  \cite{diamzon2025uncertainty}. The resulting formulation 
also yields analytical expressions for the probability density 
function and characteristic function of the network
output, together with closed-form approximations for its mean and covariance
through a combined input--parameter Jacobian.
We applied this framework to autonomous dynamical systems whose one-step
evolution map is represented by a MLP. Under successive
auto-regressive applications of the same uncertain network, the state and
network parameters become statistically dependent, even when their
perturbations are statistically independent initially. We derived recursive
equations for the state mean and covariance that explicitly account for this
dependence through the propagation of the state--parameter cross-covariance.
This cross-covariance is essential for capturing the accumulation of parameter
uncertainty over multiple time steps.
Direct propagation of a single Gaussian, however, becomes unstable over
sufficiently long forecast horizons and cannot represent the non-Gaussian
structure generated by nonlinear dynamics. To address these limitations, we
introduced the Resampled Moment Propagation, a particle-based method that combines
local analytical moment propagation with periodic resampling. Between
resampling steps, each particle accumulates state and parameter uncertainty
through the moment equations. At resampling, the resulting local Gaussian
distributions are converted into a new particle cloud and their local
covariances are reset. This construction allows us to represent evolving 
non-Gaussian state distributions while preventing the
unbounded covariance growth observed with single-Gaussian propagation. We
also examined different resampling strategies and found that the resampling
interval is the primary parameter controlling the performance of the method.
Numerical experiments on the Lorenz-63 system showed that Resampled Moment
Propagation closely reproduces the mean and covariance of the Monte Carlo
baseline through the predictability horizon, with comparable results obtained
using as few as 100 particles. Experiments on the 200-dimensional
Kuramoto--Sivashinsky system further demonstrated that the approach can be
applied to high-dimensional chaotic dynamics. At this scale, the principal
practical limitation is the memory required to store and propagate the
state--parameter cross-covariance for each particle. Neglecting this
cross-covariance substantially underestimates the propagated uncertainty,
highlighting both its importance and the need for more efficient
representations in large-scale problems.
The proposed framework provides several opportunities for further development. For instance, analytical bounds on the growth of the covariance between 
resampling steps could provide a principled
criterion for selecting the resampling interval, replacing the current
empirical tuning based on the onset of covariance growth. Similarly, the 
memory cost of the state--parameter cross-covariance scales with the product of the state dimension and the number of network parameters, motivating reduced-order or
low-rank representations for large-scale systems.
The framework can also be extended to other activation functions, with the
accuracy of the local approximation depending on their structure, and can
accommodate higher-order expansions. Exact per-layer moment-matching methods,
such as \citet{kuang2026exactmoments}, provide an alternative to linearization
for the activation functions they support. Extending our joint
input--parameter uncertainty treatment to such methods represents one
possible direction. Another is the incorporation of an explicit
measurement-update step, as in \citet{kuang2025adf}, extending the present
pure-forecasting framework toward recursive Bayesian state estimation.

\section*{Acknowledgements}
\noindent
This work was supported by the U.S. Department of Energy (DOE) under grant ``Resolution-invariant deep learning for accelerated propagation of epistemic and aleatory uncertainty in multi-scale energy storage systems, and beyond,'' contract number DE-SC0024563.

\appendix

\section{Particle Degeneracy under Resampling}
\label{app:particle_degeneration}

Consider the GMM resampling strategy in Algorithm~\ref{alg:rmp} with
$S>1$ particles and $L>1$ local samples drawn from each particle. At each
resampling step, the resulting pool contains $SL$ samples, from which $S$
samples are selected uniformly without replacement. Since the selected
samples do not necessarily contain descendants from every particle, some
particle lineages may be lost during resampling. We quantify this effect by
computing the probability that a given particle contributes no descendants
to the resampled particle cloud.
The sampling process can be described by a multivariate hypergeometric
distribution. Suppose a finite population contains $K_i$ objects of type
$i$, $i=1,\ldots,c$, with total population size
$$\displaystyle N=\sum_{i=1}^c K_i.$$ 
If $n$ objects are sampled uniformly without
replacement and $X_i$ denotes the number of sampled objects of type $i$, then
\begin{equation}
\Pr(X_1=k_1,\ldots,X_c=k_c)
=
\frac{\displaystyle\prod_{i=1}^c \binom{K_i}{k_i}}
{\displaystyle\binom{N}{n}},
\qquad
\sum_{i=1}^c k_i=n.
\end{equation}
To apply this result to our resampling procedure, fix a particle $i$ and
partition the sample pool into two groups: the $L$ local samples generated
from particle $i$, and the $L(S-1)$ samples generated from all remaining
particles. Let $X_i$ denote the number of descendants of particle $i$ among
the $S$ samples retained after resampling. The probability that particle $i$
has no descendants is therefore
\begin{equation}
\Pr(X_i=0)
=
\frac{\displaystyle \binom{L}{0}\binom{L(S-1)}{S}}
{\displaystyle\binom{LS}{S}}
=
\frac{\displaystyle\binom{L(S-1)}{S}}
{\displaystyle\binom{LS}{S}}.
\label{eq:particle_death_probability}
\end{equation}
Equivalently,
\begin{equation}
\Pr(X_i=0)
=
\prod_{j=0}^{S-1}
\frac{L(S-1)-j}{LS-j}.
\label{eq:particle_death_product}
\end{equation}
For example, when $L=10$ and $S=10$,
\begin{equation}
\Pr(X_i=0)
=
\frac{\displaystyle\binom{90}{10}}{\displaystyle\binom{100}{10}}
\approx 0.33.
\end{equation}
Thus, even for a relatively small particle cloud, a substantial fraction of
the particle lineages can be lost at a single resampling step.
The limiting behavior as $L\rightarrow\infty$ follows directly from
\eqref{eq:particle_death_product}. Dividing the numerator and denominator of
each factor by $L$ gives
\begin{equation}
\Pr(X_i=0)
=
\prod_{j=0}^{S-1}
\frac{(S-1)-j/L}{S-j/L}.
\end{equation}
For fixed $S$, $j/L\rightarrow0$ as $L\rightarrow\infty$, and hence
\begin{equation}
\lim_{L\rightarrow\infty}\Pr(X_i=0)
=
\left(1-\frac{1}{S}\right)^S.
\end{equation}
Taking $S\rightarrow\infty$ subsequently yields
\begin{equation}
\lim_{S\rightarrow\infty}
\lim_{L\rightarrow\infty}
\Pr(X_i=0)
=
\frac{1}{e}
\approx 0.368.
\end{equation}
Since all $S$ particles are statistically equivalent under this resampling
procedure, $\Pr(X_i=0)$ is also the expected fraction of particle lineages
lost at each resampling step. Thus, in the large-$S$, large-$L$ limit,
approximately $37\%$ of the existing lineages are expected to disappear at
each resampling event. Repeated resampling can therefore produce substantial
particle degeneration, consistent with the sample impoverishment observed
numerically for the Lorenz-63 system.

%\section*{References}
\bibliographystyle{plainnat}
\bibliography{main}

\end{document}